\documentclass[journal,twoside,web]{ieeecolor}
\usepackage{tmi}
\usepackage[noadjust]{cite}
\usepackage{amsmath,amssymb,amsfonts}
\usepackage{algorithmic}
\usepackage{graphicx}
\usepackage{textcomp}

\usepackage{multirow}
\usepackage{multicol}
\usepackage{makecell}
\usepackage{bm}
\usepackage{colortbl}
\usepackage[table]{xcolor}

\usepackage{hyperref}
\makeatletter
\let\NAT@parse\undefined
\makeatother
\hypersetup{hidelinks,
            colorlinks=true,
            linkcolor=blue,
            anchorcolor=blue,
            citecolor=blue,
            backref=page}

\def\BibTeX{{\rm B\kern-.05em{\sc i\kern-.025em b}\kern-.08em
    T\kern-.1667em\lower.7ex\hbox{E}\kern-.125emX}}
\begin{document}

% \title{Patient-to-Patient Multi-Grained Knowledge- Enhanced 3D Vision-Language Pre-training}

\title{MG-3D: Multi-Grained Knowledge-Enhanced \\ 3D Medical Vision-Language Pre-training\vspace{-0.3em}}

\author{Xuefeng Ni, Linshan Wu, Jiaxin Zhuang, Qiong Wang, Mingxiang Wu, Varut Vardhanabhuti, Lihai Zhang, Hanyu Gao, and Hao Chen, \IEEEmembership{Senior Member, IEEE}\vspace{-3.0em}
\thanks{This work was supported by the Hong Kong Innovation and Technology Fund (Project No. MHP/002/22 and No. GHP/006/22GD), and the Research Grants Council of the Hong Kong Special Administrative Region, China (Project No. T45-401/22-N). Xuefeng Ni, Linshan Wu, Jiaxin Zhuang, and Hao Chen are with the Department of Computer Science and Engineering, The Hong Kong University of Science and Technology, Hong Kong, China.  Qiong Wang is with the Shenzhen Institutes of Advanced Technology, Chinese Academy of Sciences, Shenzhen, China. Mingxiang Wu is with the Department of Radiology, Shenzhen People’s Hospital, Luohu, Shenzhen, China. Varut Vardhanabhuti is with the Department of Diagnostic Radiology, The University of Hong Kong, Hong Kong, China. Lihai Zhang is with the Department of Orthopaedics, Chinese PLA General Hospital, Beijing, China. Hao Chen and Hanyu Gao are with the Department of Chemical and Biological Engineering, The Hong Kong University of Science and Technology, Hong Kong, China. Hao Chen is also affiliated with the Division of Life Science, The Hong Kong University of Science and Technology, Hong Kong, China. Corresponding author: Hao Chen  (e-mail: jhc@cse.ust.hk). }}

\maketitle

\begin{abstract}
3D medical image analysis is pivotal in numerous clinical applications. However, the scarcity of labeled data and limited generalization capabilities hinder the advancement of AI-empowered models. Radiology reports are easily accessible and can serve as weakly-supervised signals. However, large-scale vision-language pre-training (VLP) remains underexplored in 3D medical image analysis. Specifically, the insufficient investigation into multi-grained radiology semantics and their correlations across patients leads to underutilization of large-scale volume-report data. 

Considering intra-patient cross-modal semantic consistency and inter-patient semantic correlations, we propose a multi-task VLP method, MG-3D, pre-trained on large-scale data (47.1K), addressing the challenges by the following two aspects: 1) Establishing the correspondence between volume semantics and multi-grained medical knowledge of each patient with cross-modal global alignment and complementary modality-guided local reconstruction, ensuring intra-patient features of different modalities cohesively represent the same semantic content; 2) Correlating inter-patient visual semantics based on fine-grained report correlations across patients, and keeping sensitivity to global individual differences via contrastive learning, enhancing the discriminative feature representation. Furthermore, we delve into the scaling law to explore potential performance improvements. Comprehensive evaluations across nine uni- and cross-modal clinical tasks are carried out to assess model efficacy. Extensive experiments on both internal and external datasets demonstrate the superior transferability, scalability, and generalization of MG-3D, showcasing its potential in advancing feature representation for 3D medical image analysis. Code will be available: https://github.com/Xuefeng-Ni/MG-3D.
\end{abstract}


\vspace{-1.0em}
\section{Introduction}
\label{sec:introduction}
\IEEEPARstart{T}{hree}-dimensional (3D) radiologic image analysis plays a crucial role in healthcare, offering detailed insights into anatomical structures and diseases of patients. AI-driven vision techniques promise to effectively assist various clinical tasks with 3D radiologic images, including disease diagnosis \cite{luna}, surgical planning \cite{swin3d}, prognosis prediction \cite{stoic}, and beyond.  

Foundation models (FMs) for 3D medical images built via large-scale pre-training are expected to advance extensive clinical tasks \cite{misfm}. In the realm of pre-training paradigms for building vision FMs, self-supervised learning has emerged as a label-efficient way to learn robust and generalizable visual feature representations, advancing diverse clinical tasks and has recently gained significant attention \cite{swinmm, swinunetr}, while it overlooks the valuable knowledge from radiologists' reports. 

\begin{figure}[!t]
\centerline{\includegraphics[width=\columnwidth]{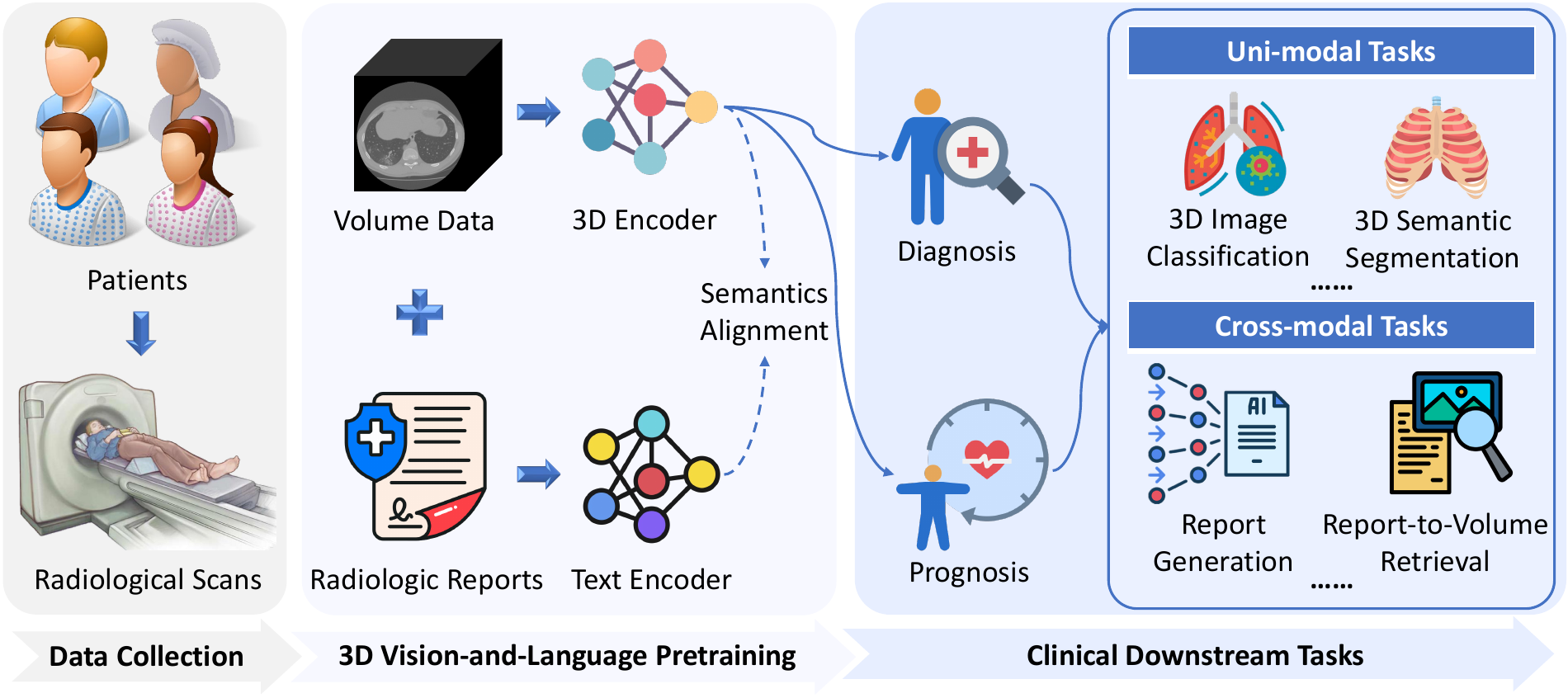}}
\setlength{\abovecaptionskip}{-1.0pt}
\caption{Overview of the 3D medical VLP framework. After collecting large-scale 3D volume-report data from diverse patient groups, the 3D vision encoder can learn radiology knowledge from reports by aligning semantic features across modalities. This framework can advance various clinical tasks, such as diagnosis, treatment, prognosis, and beyond.}
\label{Fig_Flow}
\vspace{-1.5em}
\end{figure}

Radiology reports, paired with 3D medical images, are easily available and provide highly detailed medical semantics analyses, which can serve as weakly supervised signals in VLP, strengthening the radiologic semantics capturing, as shown in Fig.~\ref{Fig_Flow}. VLP in the medical domain can be divided into 2D and 3D strategies.

2D VLP strategies, using large-scale paired X-ray-report data \cite{m3ae, ptunifier}, have demonstrated promising prospects in enhancing visual understanding and can be adapted to extensive downstream tasks, however 3D VLP has not been sufficiently explored since complex anatomical structures of 3D volumes and the verbose nature of radiology reports bring greater challenges to the alignment between visual semantics and textural descriptions. 3D VLP in medical domains also has its own typical challenges compared with that in general domains, since short-text captions suffice for describing sparse 3D point clouds \cite{point} and video sequences \cite{video} in general domains, while the rich semantics contained within lengthy reports introduce complexities. In particular, every report sentence corresponds to different semantics in a high-dimensional global 3D volume.

Despite recent advancements in 3D VLP strategies \cite{ct_clip, merlin}, how to effectively strengthen the 3D visual representation with multi-grained radiologic knowledge from reports via cross-modal interaction remains challenging. Three primary challenges limit the pre-training performance: 

\begin{itemize}
\item [a)] {\em Pre-training Data Scale and Model Capacity}:
An inadequate understanding of scaling laws for 3D medical VLP can lead to data underutilization and model bias, limiting the model efficacy and generalization abilities.

\item [b)] {\em Pre-training Strategy}:
Current 3D medical VLP methods \cite{ct_clip, t3d} overlook multi-grained correspondence between modalities within individual patients (intra-patient) and multi-grained semantic relationships across patients (inter-patient), resulting in a deficiency in mining representative multi-grained visual representations among large-scale patient groups from rich report semantics.

\item [c)] {\em Extensive Validation}:
There is a scarcity of extensive validation, across a wide range of clinical downstream tasks, for comprehensively evaluating 3D medical vision-and-language models (VLMs).
\end{itemize}

To address the challenges, we aim to propose a generalizable 3D medical VLM with a novel pre-training strategy based on large-scale volume-report data. We observe that each report provides multi-grained descriptions for various anatomies and lesions within their paired 3D volume. Moreover, different reports offer consistent fine-grained descriptions for similar visual characteristics and contrasting expressions for distinct pathological conditions within the same anatomical structures. Inspired by this, we propose MG-3D to mine intra-patient multi-grained semantics and inter-patient semantics correlations via the delicate design of multiple pretext tasks with effective cross-modal interaction. Furthermore, we scale up the pre-training data size and increase the model capacity to verify the scalability of MG-3D, leading to further performance improvement of our proposed VLM. To thoroughly evaluate the model's effectiveness and generalization ability, extensive validations across nine clinical tasks are carried out, including disease diagnosis, prognosis, organ and lesion segmentation, report generation, report-to-volume retrieval, etc. The abovementioned consideration leads to our main contributions:

\begin{itemize}
\item The proposed 3D VLMs pre-trained on the current largest-scale volume-report data consistent to CT-CLIP \cite{ct_clip} achieve superior performance through extensive validation across nine clinical tasks, which verify the effectiveness, scalability, and generalization ability of MG-3D.

\item To enrich the vision comprehension by report guidance, we propose to unify volume semantics with multi-grained medical knowledge for each patient via global cross-modal alignment and complementary modality-guided local information reconstruction.

\item To strengthen discriminative representation, we correlate visual semantics across patients based on fine-grained report correlations and keep sensitivity to global semantic differences of various patients via contrastive learning.

\item To enhance the understanding of MG-3D’s model behavior on different tasks, we delve into the scaling law \cite{scaling} of data and model capacity, providing insights into how our proposed VLMs learn and adapt as resources increase.
\end{itemize}

In the reminder of this paper, Section~\ref{sec:survey} offers a review of related work, Section~\ref{sec:method} describes the proposed contributions at length, Section~\ref{sec:experiment} evaluates the performance of the proposed VLM via comprehensive comparisons with related state-of-the-arts (SOTA), Section~\ref{sec:conclusion} concludes the main contents.

\vspace{-0.5em}
\section{Related Work} \label{sec:survey}

\subsection{3D Medical Image Pre-training}

In recent years, self-supervised learning (SSL) has emerged as the predominant label-efficient pre-training strategy for advancing 3D vision tasks in medical imaging \cite{swinmm, swinunetr, voco, vocov2}. In particular, various pretext tasks have been developed to learn visual representations from vast amounts of unlabeled data. These tasks primarily fall into four categories: 1) {\em predictive}; 2) {\em generative}; 3) {\em contrastive}; and 4) {\em mixed} strategies. 

By applying various spatial transformations, {\em predictive tasks} enable the learning of spatial information in volumetric structures, including jigsaw puzzles \cite{jigsaw}, Rubik’s Cube recovery \cite{rubik}, and relative position \cite{ssl} or rotation angle \cite{swinunetr} prediction. {\em Generative tasks} can enhance a network's ability to perceive spatial context by reconstructing masked \cite{mae, mim}, noised \cite{d2s}, or distorted regions \cite{transvw} in 3D volumes. {\em Contrastive learning} was employed to learn feature correlations in different 3D volumes, with the selection of positive and negative pairs being a central challenge. Typically, different augmented views of a single volume served as positive pairs, while volumes from other patients were considered negative pairs \cite{contrastive}. However, capturing local differences across patients is difficult due to similarities in inter-patient anatomical structures. In order to capture local contexts, overlapping sub-volumes were selected as positive pairs, and nonintersecting sub-volumes were treated as negative pairs \cite{pcrlv2}. Furthermore, geometric similarities between volumes \cite{gsvl, alice} were introduced to learn anatomical information. {\em Mixed learning} strategies \cite{pcrl, modelgen} combined multiple types of pretext tasks to elevate task complexity and enrich feature representation.  

SSL strategies have significantly advanced data-efficient 3D medical image analysis, while they overlook the valuable knowledge from radiologists.

\begin{figure*}[!t]
\vspace{-0.5em}
\centerline{\includegraphics[width=\textwidth]{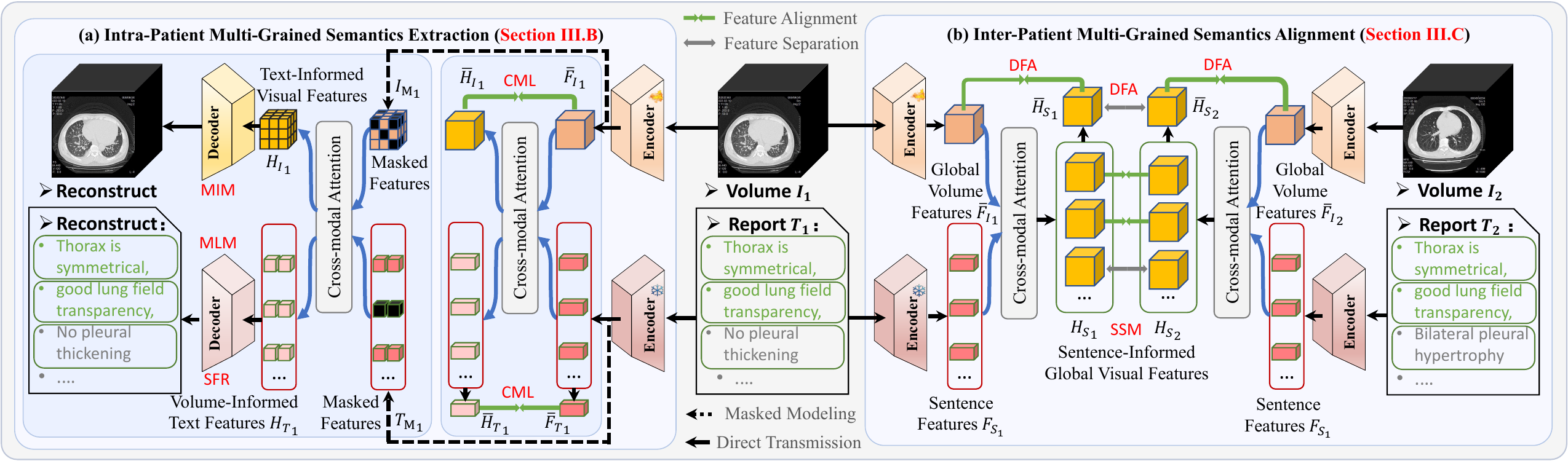}}
{
\setlength{\abovecaptionskip}{-1.0pt}
\caption{Overview of the proposed framework: (a) The left section illustrates the intra-patient multi-grained semantics extraction, consisting of cross-modal global feature alignment (CML) and complementary modality-guided local information reconstruction (MIM, MLM, and SFR). (b) The right section depicts the inter-patient multi-grained semantics alignment, generating sentence-informed global visual features for different patients, and aligning these fine-grained features (SSM) and their aggregated global features across patients (DFA) via contrastive learning. }
\label{Fig_Overview}}
\vspace{-1.5em}
\end{figure*}

\vspace{-1.0em}
\subsection{Medical Vision-Language Pre-training}

VLP has made significant progress in the 2D medical image domain, however 3D VLP has not been sufficiently researched.

In {\em 2D medical VLP}, two-lag networks \cite{clinical_bert} can model visual and textual tokens as sequences for multimodal feature fusion, using additional multimodal encoders with multi-task pre-training. However, their performance in uni- and cross-modal downstream tasks often falls short compared to two-tower structures \cite{vlp_survey} with Contrastive Language-Image Pre-Training (CLIP)-style strategies \cite{sat}, aligning visual and textual features within the same feature space to enhance cross-modal semantics understanding. Beyond CLIP-style global image-report alignment \cite{sat, CP-CLIP}, local image region-report alignment \cite{medclip} and cross-modal local reconstruction \cite{m3ae} were introduced to improve fine-grained semantic perception. In order to further analyze radiologic semantics in 2D images, additional entity descriptions \cite{mcpl} and knowledge graphs \cite{arl} were incorporated to aid VLP. Given the challenges of aligning intra-patient information from paired images and long-text reports, sentence-wise encoding \cite{prior} has been employed to align local visual features with features of split report sentences. However, other patients’ information was typically treated as negative samples in these processes. Although some similarity metrics for inter-patient global reports were proposed to better establish the inter-patient global visual correlations \cite{sat, imitate}, inter-patient multi-grained correlations were still overlooked, potentially leading to over-fitting. 

Recently, {\em 3D medical VLP} methods have attracted increasing attention. Some approaches adapted 2D or video encoders for 3D VLP \cite{radfm} or attempted to unify 2D and 3D VLP \cite{unimiss, unimedi}. However, these adaptations result in a loss of specificity in 3D visual representation due to the trade-offs between 2D and 3D tasks. To address the scarcity of text paired with 3D volumes, generative texts from Large Language Models \cite{gtgm} and web-crawled texts \cite{m3d} were used. However, the quality and comprehensiveness of these texts are not as good as that of radiology reports rich in medical knowledge. Original CLIP-style strategies with global cross-modal alignment \cite{ct_clip} have been introduced to train 3D Vision Transformers (ViT) with reports, demonstrating the effectiveness of 3D VLP in clinical tasks, however they seldom paid attention to fine-grained details related to anatomical and lesion characteristics. T3D \cite{t3d} aligned two randomly cropped global report-informed sub-volumes to advance 3D vision tasks, nevertheless it has a deficiency in correspondence between fine-grained report information and global volumes. Except for reports, several methods incorporated additional information, including paired Electronic Health Record \cite{merlin}, X-ray images \cite{biud}, and grounded organ masks \cite{ct_glip}, to assist 3D VLP. However, these strategies increase the difficulty of additional data collection.

Despite the promising prospects of 3D medical VLP, intra-patient multi-grained correspondence between modalities and inter-patient multi-grained semantics correlations have not been thoroughly researched to enhance large-scale pre-training. Moreover, it is necessary to investigate the broader contributions of 3D VLP in advancing extensive clinical tasks.

\vspace{-0.5em}
\section{Methodology}
\label{sec:method}

\subsection{Overall Framework}

Learning strategies play crucial roles in large-scale 3D medical VLP to capture radiology semantics from reports. To effectively extract the unique fine-grained characteristics of each patient, and robustly model broad patterns across diverse patients, we propose multi-task VLP strategies, enabling the network to learn generalized representations that enhance each task and foster synergistic learning across all tasks. As shown in Fig.~\ref{Fig_Overview}, our learning strategies encompass two main aspects:

% given a paired volume-report $\left\{ {{I}_{1}},{{T}_{1}} \right\}$ with other paired $\left\{ {{I}_{n}},{{T}_{n}} \right\}_{n=2}^{N}$ from $N$ patients, we separately extract their volumetric features and textual features by a shared 3D vision encoder ${{E}^{\text{I}}}$ and a frozen textual encoder ${{E}^{\text{T}}}$. The extracted features are leveraged to minimize the objective by
% %
% \begin{align}
% \label{eqn objective}
% {{\theta }^{*}},\theta _{1}^{*},...,\theta _{Z}^{*}=\underset{\theta ,{{\theta }_{1}},...,{{\theta }_{Z}}}{\mathop{\arg \min }}\,\sum\limits_{z=1}^{Z}{{{L}_{z}}\left({{D}_{{{\theta }_{z}}}}\left( E_{\theta }^{\text{I}}(I),{{E}^{\text{T}}}(T) \right) \right)}
% \end{align}
% %
% where $z$ represents the number of pretext tasks; ${{L}_{z}}$ is the loss function of the $z$-th task; ${{D}_{{{\theta }_{z}}}}$ is the decoder with parameters ${{\theta }_{z}}$ in the $z$-th task; $\theta $ denotes the parameters of the 3D vision encoder. Our learning strategies encompass two main aspects:

\paragraph{\textbf{Intra-Patient Multi-Grained Semantics Extraction}}
To strengthen the understanding of multi-grained radiological contexts within each volume-report pair, Section~\ref{sec:method}.B proposes global cross-modal alignment (CML) with a novel cross-modal interaction mechanism, modeling the global semantic correspondence between different modalities. To unify fine-grained semantic consistency and capture the complementarity of different modalities, we integrate complementary modality-guided local information reconstruction, including masked image modeling (MIM), word-level masked language modeling (MLM), and sentence-level feature reconstruction (SFR), to enhance fine-grained cross-modal knowledge transfer.
    
\paragraph{\textbf{Inter-Patient Fine-Grained Semantics Alignment}}
To ensure robust representation consistency for similar medical conditions while maintaining feature discrimination for different anatomical details across patients, Section~\ref{sec:method}.C proposes an inter-patient fine-grained semantics similarity matching (SSM) strategy to align disentangled fine-grained visual semantics among different patients. For simultaneously keeping sensitivity to global individual differences of various patients, the disentangled feature aggregation (DFA) learning is developed, differentiating globally aggregated semantics across patients and facilitating the transfer of aggregated fine-grained semantics to the 3D vision encoder.

\vspace{-1.0em}
\subsection{Intra-Patient Multi-Grained Semantics Extraction}

Given that paired volume-report data contain rich multi-grained semantic correlations, designing holistic learning strategies with efficient cross-modal interaction is crucial for data mining. Single-task VLP strategies only tend to memorize the knowledge specific to their designated tasks, leading to limited data utilization and inadequate generalization capabilities. Furthermore, cross-modal interaction in VLP plays a key role in dominant-modality information preservation and learning difficulty management. Our goal is to develop an effective multi-task learning strategy for intra-patient semantics extraction, incorporating cross-modal interaction that emphasizes the context of the dominant modality while effectively leveraging information from the complementary modality.

% \vspace{-1.0em}
% \subsection*{Complementary Modality-Guided Local Reconstruction}
% \vspace{0.4em}
% \noindent\textbf{{\em Complementary Modality-Guided Local Reconstruction}}
\subsubsection{Complementary Modality-Guided Local Reconstruction}

\begin{figure}[!t]
\vspace{-0.5em}
\centerline{\includegraphics[width=\columnwidth]{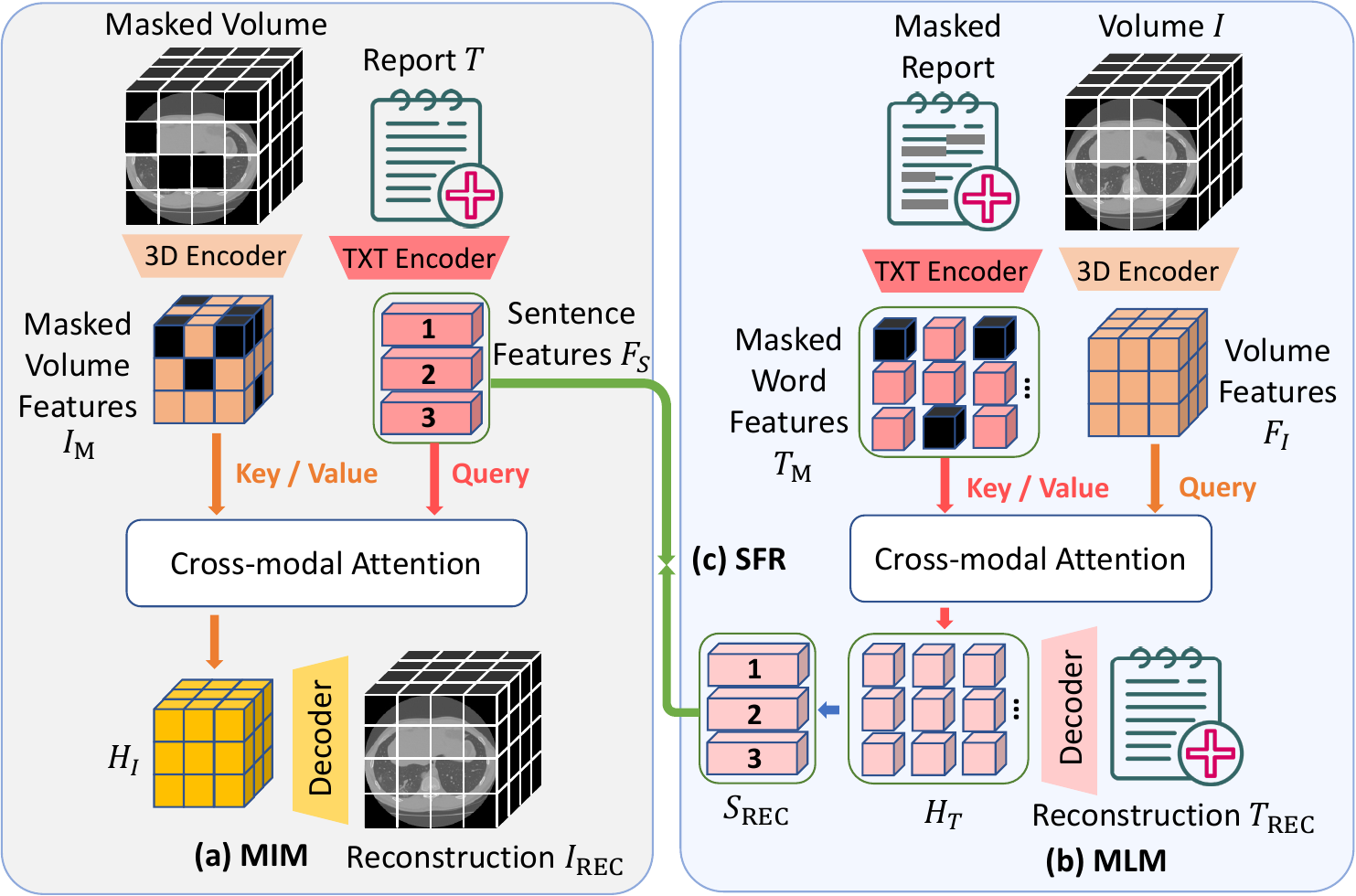}}
\setlength{\abovecaptionskip}{-1.0pt}
\caption{Complementary Modality-Guided Local Information Reconstruction: (a) With the guidance of sentence semantics, masked volumes are complemented via MIM. (b) Likewise, masked report words are reconstructed in MLM with visual assistance. (c) In SFR, the reconstructed word features are aggregated at the sentence level to ensure alignment with the original sentence features.}
\label{Fig_MIM}
\vspace{-1.5em}
\end{figure}

To ensure local context capturing with the guidance of complementary modalities, we propose to reconstruct masked sub-volumes via fine-grained report clues and recover masked reports at word- and sentence-levels with visual assistance.

% \subsubsection{Sentence-Informed Volume Reconstruction}
\paragraph{\textbf{Sentence-Informed Volume Reconstruction}}

As shown in Fig.~\ref{Fig_MIM}(a), the 3D volume is partitioned into sub-volumes and randomly masked. Compared to entire reports or brief report words, sentences provide appropriate information of medical findings, maintaining semantic coherence that supports fine-grained visual semantic analysis in 3D volumes. Thus, we aggregate word embeddings ${{F}_{T}}$, extracted from the text encoder, into sentence features ${{F}_{S}}$ by sentence-wise average pooling. Specifically, when separating a report using punctuation, we prioritize creating sentences that are as concise as possible while maintaining a complete and coherent structure. 

\begin{figure}[!t]
\vspace{-0.5em}
\centerline{\includegraphics[width=\columnwidth]{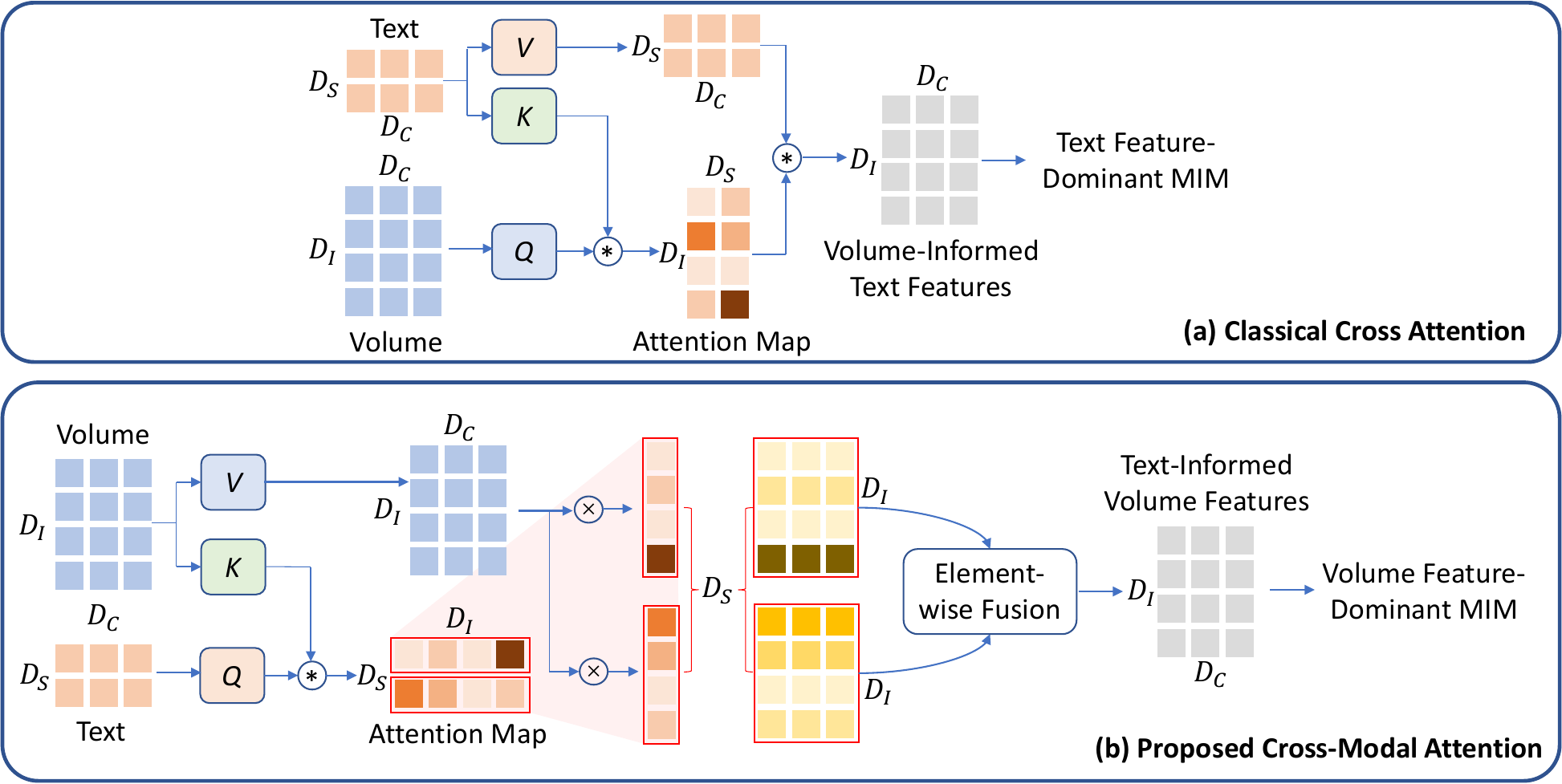}}
\setlength{\abovecaptionskip}{-1.0pt}
\caption{Cross-Modal Attention: (a) Classical mechanism:  MIM is mainly dominated by text features; (b) The proposed cross-modal attention: MIM is primarily dominated by volume features. }
\label{Fig_Coatt}
\vspace{-1.5em}
\end{figure}

The masked volume features ${{I}_{\text{M}}}$ are fused with sentence features via cross-modal attention to predict unseen sub-volumes. As shown in Fig.~\ref{Fig_Coatt}(a), MIM with classical cross attention emphasizes cross-modal generation by treating textual features as the key $K_{{{T}_{\text{M}}}}$ and value $V_{{{T}_{\text{M}}}}$ to dominate the text-to-volume generation, complicating the network modeling, since textual features might not fully capture the details in visual content, which can lead to potential information loss. To overcome this challenge, we propose a novel cross-modal attention to leverage masked volume features as the key $K_{{{I}_{\text{M}}}}$ and value $V_{{{I}_{\text{M}}}}$ to dominate MIM with the guidance of sentence semantics as the query ${{Q}_{S}}$, focusing on visual semantic context understanding. 

Specifically, as shown in Fig.~\ref{Fig_Coatt}(b), the multi-modal feature fusion between features of each sentence and unmasked sub-volume is adopted via Transformer layers, in which three types of sub-layers are integrated, including a self-attention layer, the proposed cross-modal attention layer, and a feed-forward layer. The cross-modal attention layer generates an attention map representing correlations between report sentences and local volume regions. In particular, the entire attention map is decomposed into sentence-specific attention maps in the text-modal dimension. In order to capture fine-grained volume semantics corresponding to each sentence, masked volume features ${{I}_{\text{M}}}$ are weighted by each sentence-specific attention map, leading to ${{D}_{S}}$ sentence-informed volume features ${{H}_{I}}$:

{\setlength\abovedisplayskip{-4pt}
{\small 
\begin{align}
{{H}_{I}}=\frac{\sum\nolimits_{s=0}^{{{D}_{S}}}{\text{Softmax}\left( W_{s}^{Q}{{Q}_{s}}\centerdot W_{{{I}_{\text{M}}}}^{K}K_{{{I}_{\text{M}}}}^{\top } \right)}}{\sqrt{d}\centerdot {{D}_{S}}}\centerdot W_{s}^{V}{{V}_{{{I}_{\text{M}}}}},
\end{align}}}
where $W_{s}^{Q}$, $W_{{{I}_{\text{M}}}}^{K}$, and $W_{s}^{V}$ are projection matrices of queries, keys, and values, respectively; ${{D}_{S}}$ is the sentence-level token number. $d$ is the channel dimension of the leveraged features.

A decoder with convolutional and up-sampling layers is integrated with the multi-modal fusion layers for masked sub-volume reconstruction to get the reconstructed volume ${{I}_{\text{REC}}}$. The Mean-Square-Error (MSE) loss is used as the objective:

{\setlength\abovedisplayskip{-4pt}
{\small 
\begin{align}
{{L}_{\text{MIM}}}=\sum\nolimits_{b=1}^{B}{\frac{1}{B}{{\left\| I-{{I}_{\text{REC}}} \right\|}_{2}}}.
\end{align}}}
%
% \subsubsection{Volume-Informed Report Reconstruction}
\paragraph{\textbf{Volume-Informed Report Reconstruction}}
Given report word tokens ${{T}_{\text{M}}}$ after masked modeling, we also propose the volume-guided MLM to encourage the understanding of volume-to-report correlations, as shown in Fig.~\ref{Fig_MIM}(b). Similar to the volume reconstruction process, the text-dominant multi-modal feature fusion between features of each sub-volume and remaining unmasked words ${{T}_{\text{M}}}$ is adopted via Transformer layers, leading to ${{D}_{I}}$ sub-volume-informed word features ${{H}_{T}}$:

{\setlength\abovedisplayskip{-4pt}
{\small 
\begin{align}
{{H}_{T}}=\frac{\sum\nolimits_{i=0}^{{{D}_{I}}}{\text{Softmax}\left( W_{i}^{Q}{{Q}_{i}}\centerdot W_{{{T}_{\text{M}}}}^{K}K_{{{{T}_{\text{M}}}}}^{\top } \right)}}{\sqrt{d}\centerdot {{D}_{I}}}\centerdot W_{i}^{V}{{V}_{{{T}_{\text{M}}}}},
\end{align}}}
where ${{D}_{I}}$ is the number of sub-volume tokens. The decoder for report reconstruction is designed as a simple multi-layer perceptron (MLP) with a Softmax function, resulting in the predicted masked textual entities ${{p}_{\text{MLM}}}$. The training objective is to maximize the following conditional log-likelihood:

{\setlength\abovedisplayskip{-4pt}
{\small 
\begin{align}
{{L}_{\text{MLM}}}=-\sum\nolimits_{b=1}^{B}{\sum\nolimits_{(I,{{T}_{\text{M}}})}{\frac{1}{B}\log {{p}_{\text{MLM}}}\left( {{T}_{\text{REC}}}|I,{{T}_{\text{M}}} \right)}}.
\end{align}}}
Beyond word-level reconstruction, we also develop sentence-level feature reconstruction (SFR) to enhance the hierarchical semantic context understanding, as shown in Fig.~\ref{Fig_MIM}(c). Specifically, the reconstructed word features are aggregated as ${{{S}}_{\text{REC}}}$ at the sentence level, which is further aligned with the original sentence feature representation ${{F}_{S}}$ using cosine similarity:

{\setlength\abovedisplayskip{-4pt}
{\small 
\begin{align}
\text{Sim}\left({{F}_{S}},{{S}_{\text{REC}}} \right)=\frac{{{F}_{S}}\centerdot {{S}_{\text{REC}}}}{\left\| {{F}_{S}} \right\|\centerdot \left\| {{S}_{\text{REC}}} \right\|}.
\end{align}}}
SFR loss encourages consistent sentence-level representation:

{\setlength\abovedisplayskip{-4pt}
{\small 
\begin{align}
{{L}_{\text{SFR}}}=\sum\nolimits_{b=1}^{B}{\frac{1}{B}{{\left\| \text{Sim}\left({{F}_{S}},{{S}_{\text{REC}}} \right)-\bm{1} \right\|}_{1}}},
\end{align}}}
where $\bm{1}$ is an identity matrix with the same shape as ${{S}_{\text{REC}}}$. 

% \vspace{-1.0em}
% \subsection*{Cross-Modal Global Feature Alignment}
% \vspace{0.4em}
% \noindent\textbf{{\em Cross-Modal Global Feature Alignment}}
\subsubsection{Cross-Modal Global Feature Alignment}

\begin{figure}[!t]
\vspace{-1.0em}
\centerline{\includegraphics[width=0.37\textwidth]{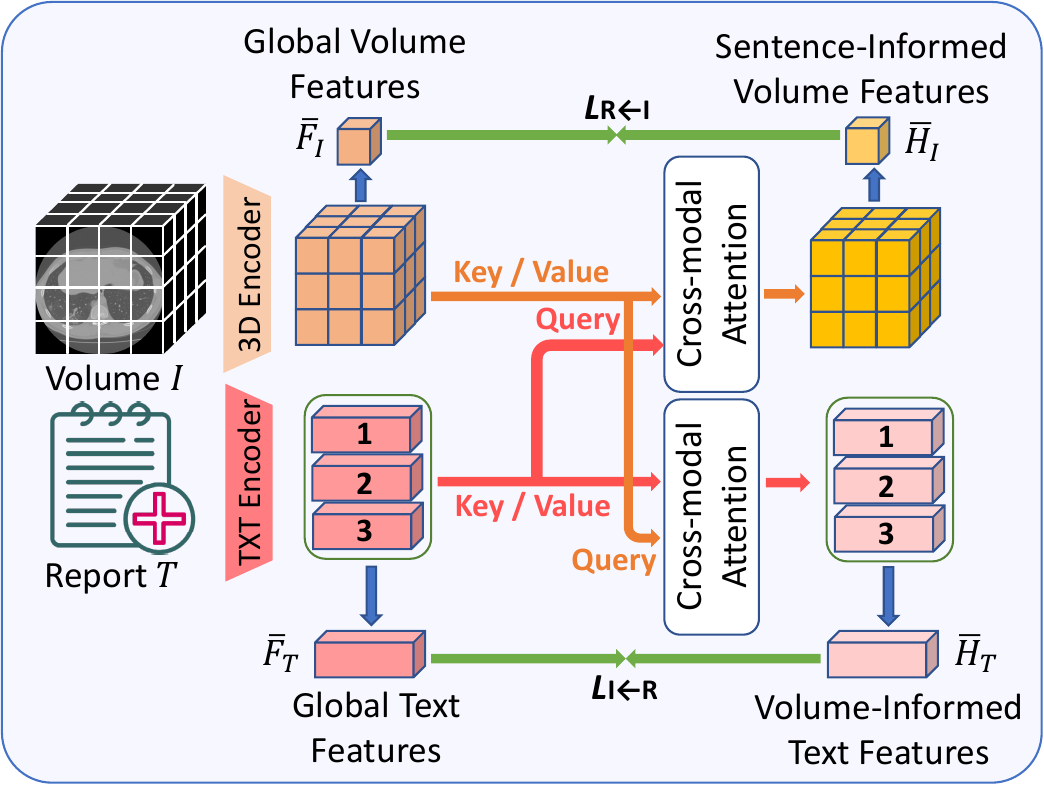}}
\setlength{\abovecaptionskip}{-1.0pt}
\caption{Intra-Patient Cross-Modal Learning: The global complementary modality-informed features are aligned with uni-modal global features to infuse cross-modal knowledge into the 3D vision encoder.}
\label{Fig_CML}
\vspace{-1.5em}
\end{figure}

The global semantics of radiology reports summarize the overall evaluation of medical conditions, providing effective clinical decision support for AI models. As shown in Fig.~\ref{Fig_CML}, global visual features ${{\bar{F}}_{I}}$ are extracted from the 3D vision encoder with average pooling, and global textual features ${{\bar{F}}_{T}}$ are aggregated from sentence features by a self-attention pooling layer. Leveraging the proposed cross-modal attention in Sec~\ref{sec:method}.B, the sentence-informed volume features and volume-informed text features are generated and further globally aggregated as ${{\bar{H}}_{I}}$ and ${{\bar{H}}_{T}}$, by the same way in uni-modal feature aggregation. To advance global visual semantics comprehension, we aim to inject global semantics into the 3D vision encoder from reports by cross-modal learning (CML), in which we maximize the normalized report-to-volume similarity between ${{\bar{F}}_{I}}$ and ${{\bar{H}}_{I}}$:

{\setlength\abovedisplayskip{-4pt}
{\small 
\begin{align}
{{L}^{\text{T}\to \text{I}}}=-\sum\nolimits_{i=1}^{B}{\log \frac{\exp \left( {{{{\bar{F}}_{I}}}^{{i}\top }}{{\bar{H}}_{I}}^{i}/\tau  \right)}{\sum\nolimits_{j=1}^{B}{\exp \left( {{{{\bar{F}}_{I}}}^{{j}\top }}{{\bar{H}}_{I}}^{i}/\tau  \right)}}},
\end{align}}}
where $\tau$ denotes the temperature parameter; $B$ is the mini-batch size. Similarly, the normalized volume-to-report similarity between ${{\bar{F}}_{T}}$ and ${{\bar{H}}_{T}}$ is optimized by

{\setlength\abovedisplayskip{-4pt}
{\small 
\begin{align}
{{L}^{\text{I}\to \text{T}}}=-\sum\nolimits_{i=1}^{B}{\log \frac{\exp \left( {{{{\bar{F}}_{T}}}^{{i}\top }}{{\bar{H}}_{T}}^{i}/\tau  \right)}{\sum\nolimits_{j=1}^{B}{\exp \left( {{{{\bar{F}}_{T}}}^{{j}\top }}{{\bar{H}}_{T}}^{i}/\tau  \right)}}}.
\end{align}}}
Thus, the CML loss ${{L}_{\text{CML}}}$ is defined as

{\setlength\abovedisplayskip{-4pt}
{\small 
\begin{align}
{{L}_{\text{CML}}}={{L}^{\text{T}\to \text{I}}}+{{L}^{\text{I}\to \text{T}}}.
\end{align}}}
Overall, the total loss for intra-patient semantics learning is

{\setlength\abovedisplayskip{-4pt}
{\small 
\begin{align}
{{L}_{\text{Intra}}}={{\lambda }_{\alpha}}({{L}_{\text{MIM}}}+{{L}_{\text{MLM}}})+{{\lambda }_{\beta}}({{L}_{\text{SFR}}}+{{L}_{\text{CML}}}).
\end{align}}}
where hyperparameters ${{\lambda }_{\alpha}}$ and ${{\lambda }_{\beta}}$ are utilized to balance the intra-patient multi-task learning. 

\vspace{-1.0em}
\subsection{Inter-Patient Multi-Grained Semantics Alignment}

\begin{figure}[!t]
\vspace{-1.0em}
\centerline{\includegraphics[width=0.48\textwidth]{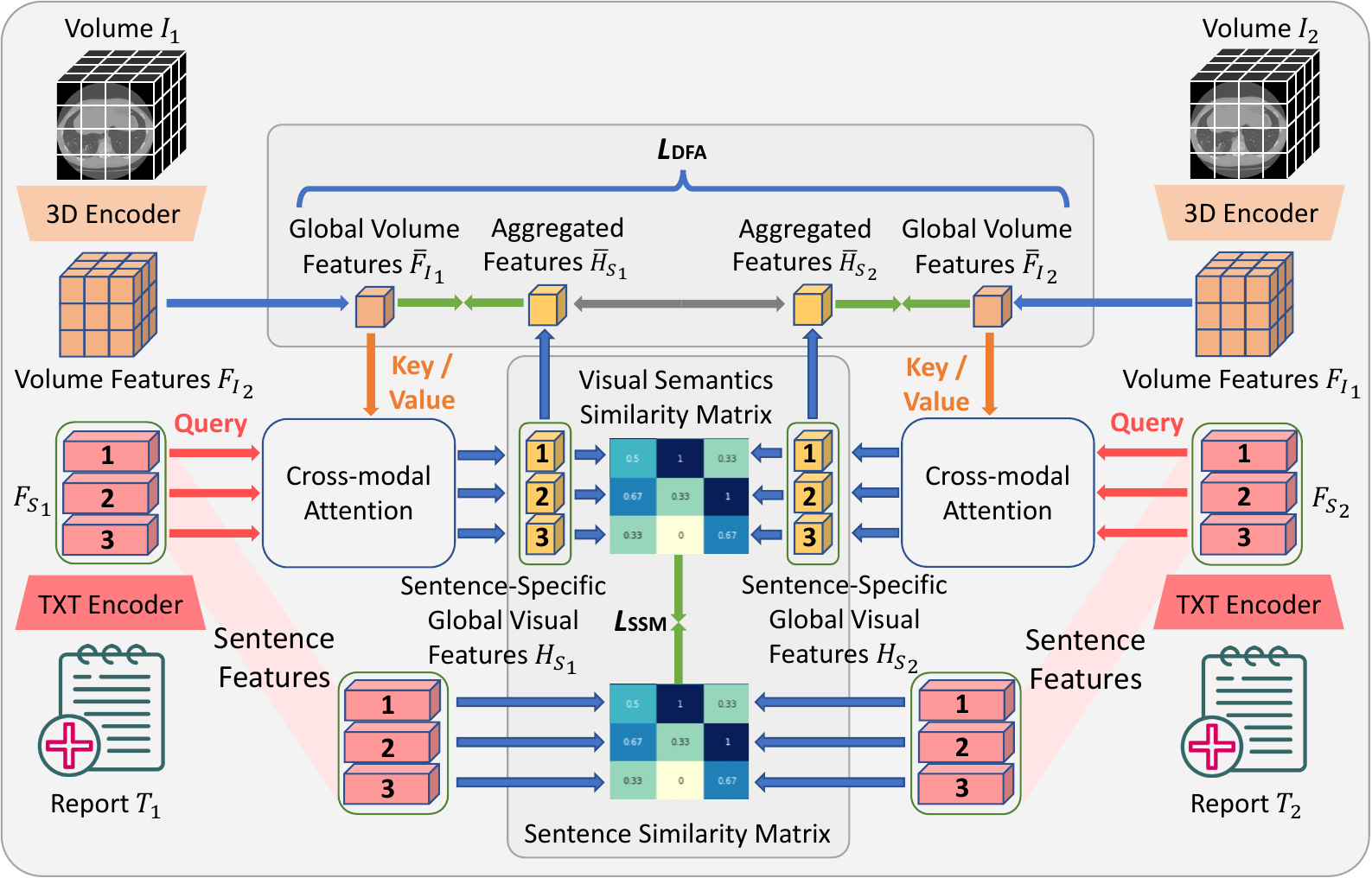}}
\setlength{\abovecaptionskip}{-1.0pt}
\caption{Inter-Patient Multi-Grained Semantics Alignment: The sentence-specific global visual features of each patient are generated by leveraging cross-modal attention between global volume features and all sentence features. From the fine-grained view, the sentence semantics similarity matrix is constructed based on sentence-level textual representations across different patients to serve as the criteria for aligning fine-grained visual semantics (SSM). From the global view, contrastive learning is implemented among the aggregated sentence-specific global visual features and global volume features across patients (DFA).}
\label{Fig_SSM}
\vspace{-1.5em}
\end{figure}

Although global individual variations can be mined by patient-wise contrastive learning to distinguish overall differences of patients’ visual semantics, the relevance of fine-grained radiology semantics across patients is also nonnegligible. Considering that different reports contain consistent fine-grained descriptions for similar visual characteristics or contrasting expressions for varying pathological conditions in the same anatomical structure, capturing inter-patient fine-grained correlations can strengthen the robustness of visual representations across large-scale patient groups. Meanwhile, we also propose to maintain the model’s sensitivity to global individual variations via disentangled feature aggregation learning.

% \vspace{-1.0em}
% \subsection*{Inter-Patient Fine-Grained Semantics Similarity Matching}
% \vspace{0.4em}
% \noindent\textbf{{\em Inter-Patient Fine-Grained Semantics Similarity Matching}}
\subsubsection{Inter-Patient Fine-Grained Semantics Similarity Matching}

We observe that different sentences in each report involve distinct properties and positions of fine-grained findings within anatomical structures, corresponding to different large-scale visual semantics of the paired volume. Inspired by this, our goal is to decouple fine-grained visual semantics with the report sentence guidance from the global volume features ${{\bar{F}}_{I}}$. 

As shown in Fig.~\ref{Fig_SSM}, in order to decouple fine-grained visual semantics, different sentence-wise features ${{F}_{S}}$ are regarded as the queries, and the global visual features ${{\bar{F}}_{I}}$ serve as the keys for implementing the cross-modal attention in Section~\ref{sec:method}.B, leading to sentence-specific global visual features:

{\setlength\abovedisplayskip{-4pt}
{\small 
\begin{align}
{{H}_{S}}=\frac{\text{Softmax}\left( W_{{{F}_{S}}}^{Q}{{Q}_{{{F}_{S}}}}\centerdot W_{{{\bar{F}}_{I}}}^{K}K_{{{\bar{F}}_{I}}}^{\top } \right)}{\sqrt{d}}\centerdot W_{{{F}_{S}}}^{V}{{V}_{{{\bar{F}}_{I}}}}.
\end{align}}
It is important to reasonably measure the relevancy between different fine-grained visual semantics. Thus, we propose to construct an inter-patient fine-grained semantics similarity matrix as the standard. Given that global report contexts of different patients are not always the same, inter-patient sentence-wise features with similar descriptions still exhibit differences if we directly send whole reports to the text encoder, leading to difficulties in precisely measuring the semantics similarity. In order to tackle this, split report sentences are separately sent into the frozen text encoder to get their sentence-wise features, ensuring consistent sentence-wise feature representation for similar fine-grained descriptions across different patients.

By calculating pair-wise feature similarities of report sentences across patients, the similarity matrix for inter-patient fine-grained semantics can be derived. We propose the semantics similarity matching loss ${{L}_{\text{SSM}}}$ to align the similarity matrix of inter-patient fine-grained visual semantics with that of inter-patient fine-grained report semantics:

{\setlength\abovedisplayskip{-4pt}
{\small 
\begin{align}{
{{L}_{\text{SSM}}}=\frac{2\sum\limits_{i=1}^{B-1}{\sum\limits_{j=i+1}^{B}{{{\left\| \text{Sim}\left( {{{F}_{S}^{i}}},{{{F}_{S}^{j}}} \right) {\tiny{-}} \text{Sim}\left( {{H}_{{{S}}}^{i}},{{H}_{{{S}}}^{j}} \right) \right\|}_{1}}}}}{B(B {\tiny{-}} 1)}.
}\end{align}}}
In mini-batch training, SSM losses between all pairs of patients are calculated once, and the average values are regressed.

% \vspace{-1.0em}
% \subsection*{Disentangled Fine-Grained Semantics Aggregation}
% \vspace{0.4em}
% \noindent\textbf{{\em Disentangled Fine-Grained Semantics Aggregation}}
\subsubsection{Disentangled Fine-Grained Semantics Aggregation}

To capture differences in inter-patient global semantics contexts, we also introduce patient-wise contrastive learning. 

Specifically, sentence-specific global visual features are aggregated as ${{\bar{H}}_{S}}$ by a self-attention pooling layer. We align the aggregated features with global visual features to model the disentangled fine-grained semantics into the vision encoder. Furthermore, the aggregated features across patients are separated for mining global individual variations, leading to the disentangled fine-grained semantics aggregation learning:

{\setlength\abovedisplayskip{-4pt}
{\small 
\begin{align}
{{L}_{\text{DFA}}}=-\sum\nolimits_{i=1}^{B}{\log \frac{\exp \left( {{{{\bar{H}}_{S}}}^{{i}\top }}{{\bar{F}}_{I}}^{i}/\tau  \right)}{\sum\nolimits_{j=1}^{B}{\exp \left( {{{{\bar{H}}_{S}}}^{{j}\top }}{{\bar{F}}_{I}}^{i}/\tau  \right)}}}.
\end{align}}}
Overall, the total loss for inter-patient semantics capturing is

{\setlength\abovedisplayskip{-4pt}
{\small 
\begin{align}
{{L}_{\text{Inter}}}={{\lambda }_{\gamma}}({{L}_{\text{SSM}}}+{{L}_{\text{DFA}}}),
\end{align}}}
where hyperparameter ${{\lambda }_{\gamma}}$ is utilized to balance the intra- and inter-patient multi-task learning. The overall training objective ${{L}_{\text{Total}}}$ of the proposed end-to-end VLP strategy is:

{\setlength\abovedisplayskip{-4pt}
{\small 
\begin{align}
{{L}_{\text{Total}}}={{L}_{\text{Intra}}}+{{L}_{\text{Inter}}}.
\end{align}}}
\vspace{-2.0em}
\section{Experiments}
\label{sec:experiment}

In this section, we first introduce the settings of pre-training and downstream tasks, after which the comparisons of our proposed pre-training strategy and VLMs against the state-of-the-arts are carried out, accompanied by ablation studies.

\vspace{-1.0em}
\subsection{Datasets and Evaluations}

\subsubsection{Pre-training Datasets}

The CTRG-Chest dataset \cite{ctrg} comprising 1,804 CT volume-report data is leveraged in ablation studies and comparisons of different pre-training strategies, especially 80\% of the volumes in this dataset are randomly selected for pre-training, while the remaining data is used for internal downstream testing. Furthermore, we adopt the largest scale data at present, the CT-RATE \cite{ct_clip} dataset with 50,188 CT volume-report pairs, for scaling up the pre-training data size and further providing more powerful VLMs compared with competitive 3D medical FMs, especially 47,149 volumes are used for pre-training, with the rest of the data allocated for internal downstream testing.

\subsubsection{Downstream Tasks and Evaluation Indexes}

\begin{table}[!b]
\vspace{-2.0em}
\scriptsize
 \setlength{\tabcolsep}{5pt}
 \renewcommand{\arraystretch}{1.3}
	\caption{Average Performance and Standard Deviations of Different Pre-training Methods in Disease Classification}
	\label{table_method_cls}
	\vspace{-0.5em}
	\centering
	\begin{tabular}{|c|c|c|c|}%{|c|l|}
%		\toprule\toprule
            \hline
		\multirow{2}{*}{\makecell[c]{Method}} & \multirow{2}{*}{\makecell[c]{Vision Encoder}} & CC-CCII & Luna16 \\
		\cline{3-4}
		& & ACC (\%) & AUC (\%) \\ \hline
  
            \multicolumn{4}{|c|}{\textbf{{\em With 3D Medical SSL}}} \\ \hline
  		MAE3D \cite{mae} & 3D ViT-B \cite{unetr} & 89.05\tiny{±2.80} & 88.26\tiny{±2.17} \\ \hline
        \rowcolor{gray!10}
        Jigsaw \cite{jigsaw} & 3D UNet \cite{ssl} & 92.89\tiny{±2.24} & 90.31\tiny{±1.12}\\ \hline
        \rowcolor{gray!10}
        Rubik++ \cite{rubik} & 3D UNet \cite{ssl} & 87.73\tiny{±4.15} & 92.39\tiny{±0.10}\\ \hline
        \rowcolor{gray!10}
        PCRLv1 \cite{pcrl} & 3D UNet \cite{ssl} & 89.80\tiny{±2.15} & 86.25\tiny{±5.39}\\ \hline
        \rowcolor{gray!10}
        PCRLv2 \cite{pcrlv2} & 3D UNet \cite{ssl} & 92.10\tiny{±0.44} & 91.96\tiny{±2.22}\\ \hline
        SwinUNETR \cite{swinunetr} & 3D Swin-B \cite{swin3d} & 91.82\tiny{±0.94} & 96.58\tiny{±0.03}\\ \hline
        SwinMM \cite{swinmm} & 3D Swin-B \cite{swin3d} & 89.51\tiny{±0.50} & 95.40\tiny{±0.98}\\ \hline
        
        \multicolumn{4}{|c|}{\textbf{{\em With 3D Medical VLP}}} \\ \hline
        M3AE \cite{m3ae} & 3D Swin-B \cite{swin3d} & 90.48\tiny{±1.02} & 94.50\tiny{±1.40}\\ \hline
        ARL \cite{arl} & 3D Swin-B \cite{swin3d} & 91.02\tiny{±0.70} & 94.57\tiny{±2.04}\\ \hline
        MRM \cite{mrm} & 3D Swin-B \cite{swin3d} & 90.88\tiny{±0.39} & 96.41\tiny{±0.10}\\ \hline
        PTunifier \cite{ptunifier} & 3D ViT-B \cite{unetr} & 72.45\tiny{±0.71} & 90.35\tiny{±1.49}\\ \hline
        CT-CLIP \cite{ct_clip} & 3D ViT-B \cite{unetr} & 79.47\tiny{±0.74} & 84.26\tiny{±0.85}\\ \hline
        M3D \cite{m3d} & 3D ViT-B \cite{swin3d} & 83.50\tiny{±1.70} & 88.65\tiny{±2.96}\\ \hline    
        \textbf{Ours} & \textbf{3D Swin-B \cite{swin3d}} & 91.90\tiny{±0.13} & \textbf{96.78\tiny{±0.06}}\\ \hline  
        \rowcolor{gray!10}
        \textbf{Ours} & \textbf{3D UNet \cite{ssl}} & \textbf{95.05\tiny{±0.58}} & 96.38\tiny{±0.37}\\ \hline  
%		\bottomrule\bottomrule
	\end{tabular}
\end{table}

\begin{table*}[!t]
\vspace{-1.5em}
\scriptsize
\setlength{\tabcolsep}{2.5pt}
 \renewcommand{\arraystretch}{1.3}
	\caption{Average Performance and Standard Deviations of Different Pre-training Methods in Organ and Lesion Segmentation (MSD Task 06 \cite{msd}, Covid-19-20 \cite{covid_19_20}, and ACDC \cite{acdc}), Prognosis Prediction (STOIC-2021 \cite{stoic}), and Cross-Modal Tasks (CTRG-Chest  \cite{ctrg} and CT-RATE \cite{ct_clip})}
	\label{table_method_seg_prog_cross}
	\vspace{-0.5em}
	\centering
	\begin{tabular}{|c|c|c|c|c|c|c|c|c|c|c|c|}%{|c|l|}
%		\toprule\toprule
            \hline
		\multirow{2}{*}{\makecell[c]{Method}} & \multirow{2}{*}{\makecell[c]{Vision Encoder}} & MSD Task 06 & Covid-19-20 & ACDC & STOIC-2021 & \multicolumn{4}{c|}{CTRG-Chest (Internal)} & \multicolumn{2}{c|}{CT-RATE (Internal)} \\
		\cline{3-12}
		& & \multicolumn{3}{c|}{Dice (\%)} & AUC (\%) & BLEU-1 & BLEU-2 & BLEU-3 & BLEU-4 & AUC (\%) & Recall@50 \\ \hline
  
            \multicolumn{12}{|c|}{\textbf{{\em With 3D Medical SSL}}} \\ \hline
  		MAE3D \cite{mae} & 3D ViT-B \cite{unetr} & 51.91\tiny{±2.55} & 62.50\tiny{±5.86} & 88.33\tiny{±0.20} & 69.48\tiny{±6.94} & 53.27 & 43.19 & 36.90 & 32.51 & 66.60 & 3.85 \\ \hline
        Jigsaw \cite{jigsaw} & 3D Swin-B \cite{swin3d} & 56.94\tiny{±1.62} & 68.37\tiny{±4.75} & 88.50\tiny{±0.44} & 72.76\tiny{±9.72} & 56.56 & 45.71 & 39.09 & 34.51 & 58.05 & 2.93 \\ \hline
        Rubik++ \cite{rubik} & 3D Swin-B \cite{swin3d} & 59.20\tiny{±2.22} & 68.85\tiny{±4.76} & 88.49\tiny{±0.50} & 71.02\tiny{±8.49} & 52.32 & 42.09 & 36.13 & 31.95 & 59.83 & 2.57 \\ \hline
        PCRLv1 \cite{pcrl} & 3D Swin-B \cite{swin3d} & 51.83\tiny{±1.56} & 68.60\tiny{±5.26} & 88.47\tiny{±0.48} & 70.69\tiny{±6.74} & 58.64 & 48.11 & 40.97 & 35.95 & 62.09 & 2.80 \\ \hline
        PCRLv2 \cite{pcrlv2} & 3D Swin-B \cite{swin3d} & 60.46\tiny{±1.48} & 68.50\tiny{±4.44} & 88.61\tiny{±0.25} & 70.98\tiny{±6.76} & 58.76 & 48.47 & 41.47 & 36.46 & 58.88 & 3.46 \\ \hline
        SwinUNETR \cite{swinunetr} & 3D Swin-B \cite{swin3d} & 59.65\tiny{±1.44} & 69.30\tiny{±4.80} & 88.49\tiny{±0.50} & 72.11\tiny{±7.66} & 55.46 & 45.03 & 38.55 & 33.94 & 63.92 & 3.69 \\ \hline
        SwinMM \cite{swinmm} & 3D Swin-B \cite{swin3d} & 58.61\tiny{±0.68} & 67.96\tiny{±5.82} & 88.21\tiny{±0.85} & 72.35\tiny{±9.98} & 58.30 & 46.46 & 39.06 & 33.82 & 61.84 & 2.40 \\ \hline
        
        \multicolumn{12}{|c|}{\textbf{{\em With 3D Medical VLP}}} \\ \hline
        M3AE \cite{m3ae} & 3D Swin-B \cite{swin3d} & 60.54\tiny{±1.30} & 69.06\tiny{±4.92} & 88.77\tiny{±0.43} & 71.89\tiny{±4.30} & 59.91 & 48.62 & 41.29 & 36.20 & 63.60 & 3.09 \\ \hline
        ARL \cite{arl} & 3D Swin-B \cite{swin3d} & 58.79\tiny{±2.96} & 68.99\tiny{±4.78} & 88.65\tiny{±0.37} & 74.55\tiny{±7.70} & 60.28 & 49.15 & 42.18 & 37.16 & 63.18 & 2.93 \\ \hline
        MRM \cite{mrm} & 3D Swin-B \cite{swin3d} & 60.40\tiny{±1.25} & 69.00\tiny{±5.37} & 88.59\tiny{±0.52} & 74.45\tiny{±5.26} & 59.17 & 47.74 & 40.08 & 35.52 & 63.61 & 2.99 \\ \hline
        PTunifier \cite{ptunifier} & 3D ViT-B \cite{unetr} & 34.31\tiny{±2.84} & 62.26\tiny{±6.30} & 87.27\tiny{±0.66} & 63.60\tiny{±2.96} & 58.31 & 47.73 & 40.34 & 35.10 & 65.09 & 2.99 \\ \hline
        CT-CLIP \cite{ct_clip} & 3D ViT-B \cite{unetr} & 22.97\tiny{±8.61} & 56.58\tiny{±6.65} & 62.92\tiny{±2.23} & 63.75\tiny{±6.49} & 59.52 & 47.95 & 40.81 & 36.12 & 57.85 & 2.27 \\ \hline
        M3D \cite{m3d} & 3D ViT-B \cite{unetr} & 45.85\tiny{±1.44} & 62.04\tiny{±5.90} & 88.41\tiny{±0.84} & 66.96\tiny{±7.77} & 55.22 & 45.39 & 38.70 & 33.81 & 62.09 & 2.80 \\ \hline
        \textbf{Ours} & \textbf{3D Swin-B \cite{swin3d}} & \textbf{62.15\tiny{±1.15}} & \textbf{70.27\tiny{±4.55}} & \textbf{89.01\tiny{±0.18}} & \textbf{76.62\tiny{±7.85}} & \textbf{62.07} & \textbf{50.16} & \textbf{42.58} & \textbf{37.17} & \textbf{67.96} & \textbf{3.88} \\ \hline  
%		\bottomrule\bottomrule
	\end{tabular}
\vspace{-2.5em}
\end{table*}

We carry out extensive validation on nine uni- and cross-modal tasks to comprehensively assess the performance of MG-3D. The downstream tasks and evaluation indexes are listed as follows.

% \begin{itemize}
{\bf Disease Classification}: The CC-CCII \cite{ccii} dataset with 4,178 volumes for pneumonia classification and LUNA16 \cite{luna} dataset for nodule classification with 888 volumes are leveraged for external testing. In particular, 2,785 volumes from CC-CCII and 623 volumes from LUNA16 are randomly selected for full fine-tuning, and the remaining data are used for testing. Accuracy (ACC) \cite{ccii} and Area Under Curve (AUC) \cite{luna} are used to evaluate the model performance.

{\bf Lesion Segmentation}: The MSD Task 06 \cite{msd} dataset for tumor segmentation with 63 volumes and the Covid-19-20 \cite{covid_19_20} dataset for pneumonia segmentation with 189 volumes are adopted for external testing, especially 51 volumes in MSD Task 06 and 151 volumes in Covid-19-20 were randomly selected for full fine-tuning, and the rest are testing data. Dice \cite{msd} is used to access the performance.

{\bf Organ Segmentation}: The ACDC \cite{acdc} dataset for MRI cardiac multi-structure segmentation with 200 volumes is adopted for external testing, especially 160 volumes were randomly selected for full fine-tuning, and the rest are testing data. Dice is used to access the performance.

{\bf Prognosis Prediction}: The STOIC 2021 \cite{stoic} dataset for pneumonia severity prediction with 2,000 volumes is utilized for external testing, in which 1,600 volumes are randomly selected for full fine-tuning, and the remaining data are used for testing. AUC is used to test the model performance.

{\bf Report Generation}: The CTRG-Chest dataset is used in this task, in which 1,443 volumes the same as those in pre-training are employed for full fine-tuning, and the rest of the data in the dataset is used for testing. We use BLEU scores to measure the quality of generated reports.

{\bf Vocabulary Prompt-Driven Anomaly Classification}: We randomly sampled 1,000 samples from the CT-RATE pre-training dataset (approximately 2\% of all the training data) for the open vocabulary fine-tuning (CT-VocabFine) \cite{ct_clip}, and the rest 3,039 samples in the dataset is used for testing the few-shot performance. AUC is used to assess the performance.

{\bf Report-to-Volume Retrieval}: The dataset splitting is the same as that in the Vocabulary Prompt-Driven Anomaly Classification. Recall \cite{ct_clip} is used to evaluate the performance.
% \end{itemize}

\vspace{-1.0em}
\subsection{Implementation Details}

\subsubsection{Pre-training Setup}

For the implementation, the 3D Swin Transformer \cite{swin3d} with a hierarchical structure serves as the default 3D vision encoder, and the RadBERT \cite{radbert} is adopted as the text encoder. The text encoder's parameters are frozen to force the 3D vision encoder to learn radiology semantics. The input volume undergoes center-cropping with cropping ratios of [0.88, 0.66, 0.88] to increase the foreground proportion and then is resized to the size of [128, 96, 128]. The batch size per GPU is 4. Learning rates for the 3D vision encoder and the multi-modal fusion module are separately set to 2e-5 and 1e-4. The VLM is optimized by an AdamW optimizer with a weight decay of 0.01 for 300,000 steps in CTRG-Chest and 100,000 steps in CT-RATE, respectively. On the basis of empirical tests, the coefficients [${{\lambda }_{\alpha}}$, ${{\lambda }_{\beta}}$, ${{\lambda }_{\gamma}}$] for balancing the losses are set to [1.0, 0.1, 0.1]. MG-3D was implemented by PyTorch and MONAI, with the pre-training conducted on 4 H800 GPUs.

% The maximum input length for text tokens is set to 250. The number of the proposed cross-attention layers in the multi-modal fusion module is configured to 2, with the head number set to 12, and a hidden state dimension of 768. The mask ratio for MIM is set to 50\%. A warm-up period of 10,000 steps is implemented. 

\subsubsection{Downstream Settings}

The configuration settings for each downstream task are summarized as follows. For disease classification and prognosis prediction, a linear classifier is introduced as the decoder. The decoder of Swin-UNETR \cite{swin3d} is introduced for organ and lesion segmentation. In report generation, a knowledge-enhanced report generator \cite{m2kt} with a transformer layer is introduced. In the CT-VocabFine, abnormality probabilities are calculated by measuring the similarity between the visual features and abnormal text features. For report-to-volume retrieval, the cosine similarity between the global volume features and the global report features is measured to return the top $K$ volumes for a given query report. An H800 or 3090 GPU was used for downstream fine-tuning.

\vspace{-1.0em}
\subsection{Comparisons with Medical VLP and 3D SSL Methods}

We compare MG-3D with 7 competitive 3D medical SSL methods, such as generation-based MAE3D \cite{mae} and SwinMM \cite{swinmm}, prediction-based Jigsaw \cite{jigsaw} and Rubik++ \cite{rubik}, multi-task learning-based PCRLv1 \cite{pcrl}, PCRLv2\cite{pcrlv2}, and SwinUNETR \cite{swinunetr}, and 6 recent medical VLP methods, including local reconstruction-based M3AE \cite{m3ae} and MRM \cite{mrm}, knowledge-enhanced ARL \cite{arl}, CLIP-like CT-CLIP \cite{ct_clip} and M3D \cite{m3d}, and multi-task learning-based PTunifier \cite{ptunifier}. Notably, M3AE, MRM, and ARL were originally designed for 2D medical VLP, for which we adapted these methods for our 3D scenario by substituting the vision encoder with 3D ones. To achieve a fair comparison, we reproduced the aforementioned methods by maintaining a consistent pre-training dataset, CTRG-Chest.  

\subsubsection{Disease Classification}

% Global semantics understanding is essential for effective decision-making in the classification of lung nodules and pneumonia.
Table~\ref{table_method_cls} presents the average performance of three bootstrapping iterations on two external datasets. The results show that 3D UNet, with MG-3D, has significant superiority in differentiating pneumonia types, validating the efficacy of incorporating global report semantics into pre-training for improved diagnosis. When the 3D vision encoder is configured as the 3D Swin Transformer-Base (Swin-B), MG-3D consistently outperforms all related methods. Identifying small lung nodules within large-scale volumes poses a significant challenge to some comparative pre-training methods; in contrast, MG-3D with superior generalization ability successfully meets this challenge.

\subsubsection{Lesion Segmentation}

The 3rd and 4th columns of Table~\ref{table_method_seg_prog_cross} show the average performance of three bootstrapping iterations on two external datasets. The decoders from Swin-UNETR \cite{swinunetr} and UNETR \cite{unetr} are separately incorporated for 3D Swin-B and 3D ViT-B. The results illustrate that MG-3D achieves the highest performance in tumor segmentation and pneumonia segmentation, verifying the effectiveness of patient-to-patient fine-grained radiology semantics extraction.

\subsubsection{Organ Segmentation}
Table~\ref{table_method_seg_prog_cross}'s 5th column displays the average performance of three bootstrapping iterations on the external dataset, ACDC, demonstrating that although our proposed VLM only pre-trained on CT data, it still owns superior transferability to unseen MRI image analysis tasks, verifying MG-3D's adaptability to different 3D imaging modalities. 

\subsubsection{Prognosis Prediction}

Table~\ref{table_method_seg_prog_cross}'s 6th column presents the average performance on STOIC-2021 of three bootstrapping iterations, showing that VLP strategies with 3D Swin-B outperform SSL ones, since the detailed report descriptions for lesions facilitate differentiating disease development stages.

% \begin{table}[!t]
% \setlength{\tabcolsep}{4pt}
%  \renewcommand{\arraystretch}{1.3}
% 	\caption{Prognosis Prediction Performance of Different Pre-training Methods on STOIC-2021}
% 	\label{table_method_prog}
% 	\vspace{-0.5em}
% 	\centering
% 	\begin{tabular}{|c|c|c|}%{|c|l|}
% %		\toprule\toprule
%             \hline
% 		Method & Vision Encoder & AUC (\%) \\ \hline
%             \multicolumn{3}{|c|}{\textbf{{\em With 3D medical SSL}}} \\ \hline
%   		MAE3D & 3D ViT-B & 69.48 \\ \hline
%         Jigsaw & 3D Swin-B & 72.76\\ \hline
%         Rubik++ & 3D Swin-B & 71.02\\ \hline
%         PCRLv1 & 3D Swin-B & 70.69\\ \hline
%         PCRLv2 & 3D Swin-B & 70.98\\ \hline
%         SwinUNETR & 3D Swin-B & 72.11\\ \hline
%         SwinMM & 3D Swin-B & 72.35\\ \hline
        
%         \multicolumn{3}{|c|}{\textbf{{\em With Medical VLP}}} \\ \hline
%         M3AE & 3D Swin-B & 71.89\\ \hline
%         ARL & 3D Swin-B & 74.55\\ \hline
%         MRM & 3D Swin-B & 74.45\\ \hline
%         PTunifier & 3D ViT-B & 63.60\\ \hline
%         CT-CLIP & 3D ViT-B & 63.75\\ \hline
%         M3D & 3D ViT-B & 63.96\\ \hline        
%         \textbf{Ours} & \textbf{3D Swin-B} & \textbf{76.62}\\ \hline  
% %		\bottomrule\bottomrule
% 	\end{tabular}
% \end{table}

\subsubsection{Report Generation}

% Robust fine-grained cross-modal semantics understanding is essential for this task. 
The 7th-10th columns of Table~\ref{table_method_seg_prog_cross} present the BLEU scores \cite{ctrg} on the internal dataset, indicating that VLP can mainly achieve better performance than SSL strategies owing to the medical knowledge acquired from the cross-modal learning. Furthermore, MG-3D exhibits significant performance advantages over all the related VLP methods, which is attributed to the effective fine-grained semantics disentanglement and multi-grained cross-modal alignment.

\begin{table}[!t]
\vspace{-1.0em}
\scriptsize
\setlength{\tabcolsep}{3.8pt}
 \renewcommand{\arraystretch}{1.3}
	\caption{Performance   with Different Intra-Patient Learning Strategies on MSD Task 06 \cite{msd}}
	\label{table_intra}
	\vspace{-0.5em}
	\centering
	\begin{tabular}{|c|c|c|c|c|c|c|}%{|c|l|}
%		\toprule\toprule
            \hline
		\multicolumn{3}{|c|}{Intra-Patient} & \multicolumn{2}{c|}{Cross-Modal Attention} & \multirow{2}{*}{\makecell[c]{Inter- \\ Patient}} & \multirow{2}{*}{Dice (\%)}\\
		\cline{1-5}
		MIM & MLM & CML & MIM & MLM & &\\ \hline
  		--- & --- & --- & --- & --- & --- & 54.88\\ \cline{1-7}
        Word Guidance & \checkmark & \checkmark & Classical & Classical & --- & 57.55\\ \cline{1-7}
        Sentence Guidance & \checkmark & \checkmark & Classical & Classical & --- & 59.12\\ \cline{1-7}
        Sentence Guidance & \checkmark & \checkmark & Ours & Classical & --- & 60.63\\ \cline{1-7}
        Sentence Guidance & \checkmark & --- & Ours & Ours & --- & 61.20\\  \cline{1-7}
        \textbf{Sentence Guidance} & \textbf{\checkmark} & \textbf{\checkmark} & \textbf{Ours} & \textbf{Ours} & \textbf{---} & \textbf{61.56} \\ \hline%		\bottomrule\bottomrule
	\end{tabular}
\vspace{-1.0em}
\end{table}

\subsubsection{Vocabulary Prompt-Driven Anomaly Classification \& Report-to-Volume Retrieval}

% Fine-grained and global semantics understanding are separately required in these two cross-modal tasks. 
The last two columns of Table~\ref{table_method_seg_prog_cross} show the results on the default CT-RATE test set \cite{ct_clip}, demonstrating that either SSL or VLP strategies with reconstruction-based pretext tasks yield better performance in anomaly classification, since the local reconstruction can advance fine-grained anomaly identification. The report-to-volume poses significant challenges, as the anatomical structures and distributions among different patients often exhibit similarities, leading to overlapping fine-grained descriptions across various reports. Despite the challenges, MG-3D still achieves the best performance across cross-modal tasks, which further proves its enhanced cross-modal understanding abilities.

\begin{table}[!t]
\scriptsize
\setlength{\tabcolsep}{4pt}
 \renewcommand{\arraystretch}{1.3}
	\caption{Performance with Different Losses on MSD Task 06 \cite{msd}}
	\label{table_sfr_inter}
	\vspace{-0.5em}
	\centering
	\begin{tabular}{|c|c|c|c|}%{|c|l|}
%		\toprule\toprule
            \hline
	Intra-Patient & \multicolumn{2}{c|}{Inter-Patient} & \multirow{2}{*}{Dice (\%)} \\ \cline{1-3}
  	${{L}_{\text{SFR}}}$ & ${{L}_{\text{DFA}}}$ & ${{L}_{\text{SSM}}}$ & \\ \hline
    --- & \checkmark & \checkmark & 61.54 \\ \hline
    \checkmark & --- & \checkmark & 62.11 \\ \hline
    \checkmark & \checkmark & --- & 62.22 \\ \hline
    \textbf{\checkmark} & \textbf{\checkmark} & \textbf{\checkmark} & \textbf{63.40} \\ \hline
%		\bottomrule\bottomrule
	\end{tabular}
\vspace{-2.0em}
\end{table}

\vspace{-1.0em}
\subsection{Ablation Study}

\subsubsection{Intra-Patient Learning}

Table~\ref{table_intra} illustrates the performance with different intra-patient learning strategies, tested on MSD Task 06 once with the default fine-tuning dataset settings. The vision encoder is 3D Swin-B. A comparison between the results in the 1st and 2nd rows reveals that VLP with cross-modal reconstruction enhances downstream performance compared with training from scratch. The comparison between the 2nd and the 3rd rows indicates that sentence-informed volume reconstruction outperforms word-informed reconstruction, owing to more coherent semantics and more contextual information embedded in report sentences. The last three rows verify that integrating the proposed cross-attention into MIM and MLM effectively mitigates the learning difficulty, thereby improving performance. Additionally, the global CML further strengthens the volume semantic understanding.

We further investigate the effects of the SFR loss. As shown in the 1st and 4th rows of Table~\ref{table_sfr_inter}, pre-training with SFR loss results in a significant performance enhancement, due to the enriched feature representation facilitated by hierarchical semantic context reconstruction.

% \begin{table}[!t]
% \scriptsize
% \setlength{\tabcolsep}{4pt}
%  \renewcommand{\arraystretch}{1.3}
% 	\caption{Performance with Different Losses on MSD Task 06 \cite{msd}}
% 	\label{table_sfr_inter}
% 	\vspace{-0.5em}
% 	\centering
% 	\begin{tabular}{|c|c|c|}%{|c|l|}
% %		\toprule\toprule
%             \hline
% 	Intra-Patient & Inter-Patient & Dice (\%) \\ \hline
%   	w/o ${{L}_{\text{SFR}}}$ & w ${{L}_{\text{DFA}}}$ \& ${{L}_{\text{SSM}}}$ & 61.54 \\ \hline
%   	\textbf{\multirow{3}{*}{w ${{L}_{\text{SFR}}}$}} & w/o ${{L}_{\text{DFA}}}$ & 62.11 \\ \cline{2-3}
%   	 & w/o ${{L}_{\text{SSM}}}$ & 62.22 \\ \cline{2-3}
%          & \textbf{w ${{L}_{\text{DFA}}}$} \& \textbf{${{L}_{\text{SSM}}}$} & \textbf{63.40} \\ \hline
% %		\bottomrule\bottomrule
% 	\end{tabular}
% \vspace{-1.0em}
% \end{table}

\subsubsection{Inter-Patient Learning}

To evaluate the impact of inter-patient learning, Table~\ref{table_sfr_inter} exhibits the performance without DFA loss or SSM loss, indicating that the two losses play crucial roles in the pre-training, and the inter-patient semantics alignment significantly enhances feature representativeness.

\subsubsection{Scaling Up the Pre-training Data Size}
We investigate the data scaling law \cite{scaling} of MG-3D in Table~\ref{table_data_scale}, especially we pre-trained the 3D Swin-B with varying data scales from the CT-RATE dataset and evaluated the performance in multi-anomaly classification. As anticipated, it is clear that larger pre-training data scales correlate with improved performance, verifying the scalability of MG-3D. Furthermore, Table~\ref{table_model_cap} shows that when adopting a larger scale dataset, CT-RATE (47.1K), the performance enhanced across almost all downstream tasks compared with adopting CTRG-Chest (1.4K), which further proves the importance of pre-training with large-scale data.

\begin{table}[!t]
\vspace{-1.0em}
\scriptsize
\setlength{\tabcolsep}{4pt}
 \renewcommand{\arraystretch}{1.3}
	\caption{Performance of Prompt-Driven Anomaly Classification on CT-RATE  \cite{ct_clip} with Different Pre-training Data Scale}
	\label{table_data_scale}
	\vspace{-0.5em}
	\centering
	\begin{tabular}{|c|c|c|c|c|}%{|c|l|}
%		\toprule\toprule
            \hline
		\multirow{2}{*}{\makecell[c]{Metric}} & \multicolumn{4}{c|}{Data Scale}\\
		\cline{2-5}
		& 1.4K & 2.8K & 10K & \textbf{47.1K} \\ \hline
        AUC (\%) & 64.82 & 65.12 & 67.57 & \textbf{68.62} \\ \hline
%		\bottomrule\bottomrule
	\end{tabular}
\vspace{-1.0em}
\end{table}

\begin{table*}[!t]
\vspace{-1.0em}
\scriptsize
\setlength{\tabcolsep}{2.3pt}
 \renewcommand{\arraystretch}{1.3}
	\caption{Average Performance and Standard Deviations with Different Pre-training Data Scale and Model Capacity in Nine Tasks. The Best and the Second Best Results are Separately Highlighted in Bold and Underlining.}
	\label{table_model_cap}
	\vspace{-0.5em}
	\centering
	\begin{tabular}{|c|c|c|c|c|c|c|c|c|c|c|c|c|c|c|}%{|c|l|}
%		\toprule\toprule
            \hline
		\multirow{2}{*}{\makecell[c]{Pre-training \\ Dataset}} & \multirow{2}{*}{\makecell[c]{Data \\ Size}} & \multirow{2}{*}{\makecell[c]{Vision \\ Encoder}} & CC-CCII & Luna16 & MSD Task 06 & Covid-19-20 & ACDC & STOIC-2021 & \multicolumn{4}{c|}{CTRG-Chest  (Internal)} & \multicolumn{2}{c|}{CT-RATE  (Internal)}\\
		\cline{4-15}
		&&& ACC (\%) & AUC (\%) & \multicolumn{3}{c|}{Dice (\%)} & AUC (\%) & BLEU-1 & BLEU-2  & BLEU-3 & BLEU-4 & AUC (\%) & Recall@50 \\ \hline
        CTRG-Chest & 1.4K & 3D Swin-B & 91.90\tiny{±0.13} & 96.78\tiny{±0.06} & 62.15\tiny{±1.15} & 70.27\tiny{±4.55} & 89.01\tiny{±0.18} & \underline{76.62\tiny{±7.85}} & 62.07 & 50.16 & 42.58 & 37.17 & 67.96 & \underline{3.88} \\ \hline
        \textbf{CT-RATE} & \textbf{47.1K} & 3D Swin-B & \underline{93.26\tiny{±0.67}} & \underline{96.91\tiny{±0.06}} & \underline{62.74\tiny{±0.43}} & \textbf{70.94\tiny{±4.17}} & \underline{89.09\tiny{±0.23}} & 76.31\tiny{±7.26} & \textbf{63.54} & \textbf{51.61} & \textbf{43.79} & \textbf{38.19} & \textbf{68.62} & \textbf{4.05} \\ \hline
            \textbf{CT-RATE} & \textbf{47.1K} & \textbf{3D Swin-L} & \textbf{93.61\tiny{±0.99}} & \textbf{98.20\tiny{±0.08}} & \textbf{65.52\tiny{±1.87}} & \underline{70.79\tiny{±4.06}} & \textbf{89.38\tiny{±0.41}} & \textbf{77.02\tiny{±7.27}} & \underline{62.72} & \underline{51.45} & \underline{43.63} & \underline{37.93} & \underline{68.47} & 3.06 \\ \hline
%		\bottomrule\bottomrule
	\end{tabular}
\vspace{-2.5em}
\end{table*}

% \begin{table}[!t]
% \scriptsize
% \setlength{\tabcolsep}{4pt}
%  \renewcommand{\arraystretch}{1.3}
% 	\caption{Disease Classification Performance of Different 3D Medical FMs on CC-CCII and Luna16}
% 	\label{table_fm_cls}
% 	\vspace{-0.5em}
% 	\centering
% 	\begin{tabular}{|c|c|c|c|}%{|c|l|}
% %		\toprule\toprule
%             \hline
% 		\multirow{2}{*}{\makecell[c]{Method}} & \multirow{2}{*}{\makecell[c]{Vision Encoder}} & CC-CCII & Luna16 \\
% 		\cline{3-4}
% 		& & ACC (\%) & AUC (\%) \\ \hline
  
%             \multicolumn{4}{|c|}{\textbf{{\em With 3D Medical SSL}}} \\ \hline
%   		PCRLv2 & 3D UNet & 91.00 & 85.00 \\ \hline
%         Vox2Vec & 3D FPN & 93.15 & 98.81\\ \hline
%         TransVW & 3D UNet & 91.28 & 95.11\\ \hline
%         SwinUNETR & 3D Swin-B & 91.88 & 94.22\\ \hline
%         % SwinMM & 3D Swin-B & 94.73 & 95.92\\ \hline
        
%         \multicolumn{4}{|c|}{\textbf{{\em With Medical VLP}}} \\ \hline
%         CT-CLIP & 3D ViT-B & 79.13 & 86.96\\ \hline
%         M3D & 3D ViT-B & 88.03 & 91.32\\ \hline
%         RadFM & 3D ViT-B & 87.77 & 87.91\\ \hline      
%         % \textbf{Ours} & \textbf{3D UNet} & \textbf{95.05} & 97.35\\ \hline  
%         \textbf{Ours} & \textbf{3D Swin-L} & \textbf{93.61} & \textbf{99.18} \\ \hline  
% %		\bottomrule\bottomrule
% 	\end{tabular}
%  \vspace{-2.0em}
% \end{table}

\begin{figure}[!t]
\centerline{\includegraphics[width=\columnwidth]{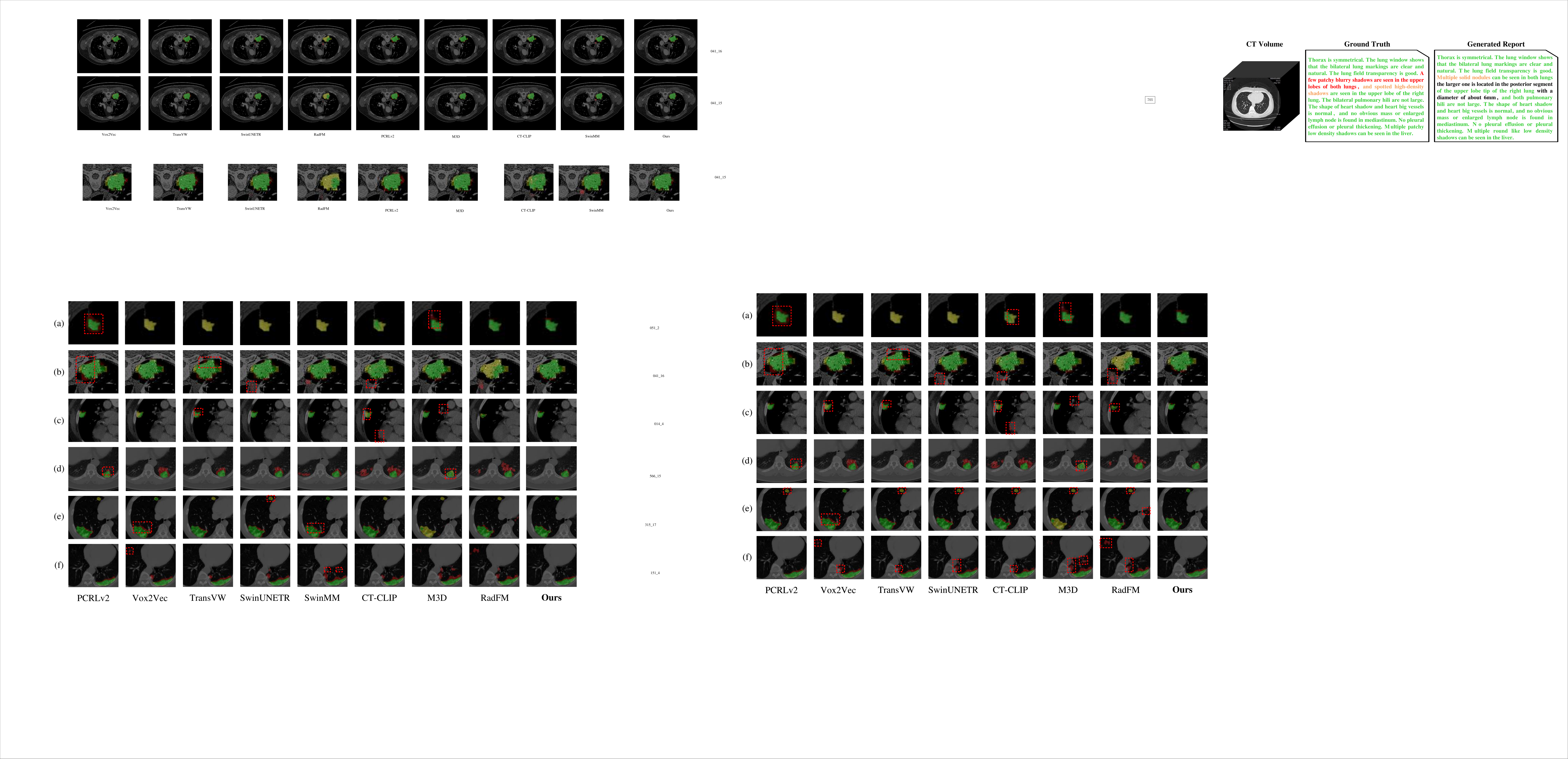}}
\setlength{\abovecaptionskip}{-1.0pt}
\caption{Qualitative results of lesion segmentation, in which regions in green, red, and yellow represent correct, wrong, and missing segmentation, respectively. The red dashed boxes highlight small errors and missing areas that require further attention.}
\label{Fig_Seg}
\vspace{-1.5em}
\end{figure}

% \begin{table}[!t]
% \scriptsize
% \setlength{\tabcolsep}{0.9pt}
%  \renewcommand{\arraystretch}{1.3}
% 	\caption{Performance on Vision Tasks of Different 3D Medical FMs}
% 	\label{table_fm_vision}
% 	\vspace{-0.5em}
% 	\centering
% 	\begin{tabular}{|c|c|c|c|c|c|c|}%{|c|l|}
% %		\toprule\toprule
%             \hline
% 		\multirow{2}{*}{\makecell[c]{Method}} & \multirow{2}{*}{\makecell[c]{Vision \\ Encoder}} & CC-CCII & Luna16 & MSD Task 06 & Covid-19-20 & STOIC-2021 \\
% 		\cline{3-7}
% 		& & ACC (\%) & AUC (\%) &  \multicolumn{2}{c|}{Dice (\%)} & AUC (\%) \\ \hline
  
%             \multicolumn{7}{|c|}{\textbf{{\em With 3D Medical SSL}}} \\ \hline
%   		PCRLv2 & 3D UNet & 91.00 & 85.00 & 58.01 & 69.45 & 73.88 \\ \hline
%         Vox2Vec & 3D FPN & 93.15 & 98.81 & 63.33 & 69.77 & 75.27 \\ \hline
%         TransVW & 3D UNet & 91.28 & 95.11 & 51.01 & 69.73 & 56.23 \\ \hline
%         SwinUNETR & 3D Swin-B & 91.88 & 94.22 & 52.83 & 65.84 & 74.46 \\ \hline
%         % SwinMM & 3D Swin-B & 61.23 & 67.20 & 71.97 \\ \hline
        
%         \multicolumn{7}{|c|}{\textbf{{\em With Medical VLP}}} \\ \hline
%         CT-CLIP & 3D ViT-B & 79.13 & 86.96 & 24.93 & 57.16 & 63.12 \\ \hline
%         M3D & 3D ViT-B & 88.03 & 91.32 & 45.07 & 61.39 & 62.71 \\ \hline
%         RadFM & 3D ViT-B & 87.77 & 87.91 & 25.70 & 57.26 & 64.85 \\ \hline
%         \textbf{Ours} & \textbf{3D Swin-L} & \textbf{93.61} & \textbf{99.18} & \textbf{65.07} & \textbf{70.79} & \textbf{77.02} \\ \hline  
% %		\bottomrule\bottomrule
% 	\end{tabular}
% \vspace{-2.0em}
% \end{table}

\begin{table}[!t]
\scriptsize
\setlength{\tabcolsep}{1.0pt}
 \renewcommand{\arraystretch}{1.3}
	\caption{Average Performance and Standard Deviations of Different 3D Medical FMs in Vision Tasks}
	\label{table_fm_vision}
	\vspace{-0.5em}
	\centering
	\begin{tabular}{|c|c|c|c|c|c|c|}%{|c|l|}
%		\toprule\toprule
            \hline
		\multirow{2}{*}{\makecell[c]{Method}} & CC-CCII & Luna16 & MSD Task 06 & Covid-19-20 & ACDC & STOIC-2021 \\
		\cline{2-7}
		& ACC (\%) & AUC (\%) &  \multicolumn{3}{c|}{Dice (\%)} & AUC (\%) \\ \hline
  
            \multicolumn{7}{|c|}{\textbf{{\em With 3D Medical SSL}}} \\ \hline
  		PCRLv2 & 91.00\tiny{±1.92} & 87.89\tiny{±1.34} & 58.01\tiny{±5.37} & 69.45\tiny{±4.96} & 89.77\tiny{±0.63}& 74.62\tiny{±10.57} \\ \hline
        Vox2Vec & 93.15\tiny{±1.54} & 98.04\tiny{±0.23} & 63.33\tiny{±3.47} & 69.77\tiny{±4.48} & 88.47\tiny{±0.56}& 75.27\tiny{±6.74} \\ \hline
        TransVW & 91.28\tiny{±1.39} & 95.55\tiny{±2.32} & 51.01\tiny{±3.81} & 69.73\tiny{±7.60} & 89.85\tiny{±0.55}& 56.23\tiny{±5.63} \\ \hline
        SwinUNETR & 91.88\tiny{±1.20} & 94.16\tiny{±1.03} & 52.83\tiny{±3.94} & 65.84\tiny{±2.86} & 88.46\tiny{±0.56}& 74.46\tiny{±7.25} \\ \hline
        % SwinMM & 3D Swin-B & 61.23 & 67.20 & 71.97 \\ \hline
        
        \multicolumn{7}{|c|}{\textbf{{\em With 3D Medical VLP}}} \\ \hline
        CT-CLIP & 79.13\tiny{±0.42} & 85.73\tiny{±0.56} & 24.93\tiny{±6.97} & 57.16\tiny{±7.24} & 88.90\tiny{±0.40}& 63.12\tiny{±4.27} \\ \hline
        M3D & 88.03\tiny{±3.83} & 90.22\tiny{±0.16} & 45.07\tiny{±1.55} & 61.39\tiny{±4.29} & 88.25\tiny{±0.72}& 62.71\tiny{±7.11} \\ \hline
        RadFM & 87.77\tiny{±0.65} & 85.69\tiny{±0.12} & 25.70\tiny{±4.46} & 57.26\tiny{±4.55} & \textbf{89.93\tiny{±0.48}} & 64.85\tiny{±7.25} \\ \hline
        \textbf{Ours} & \textbf{93.61\tiny{±0.99}} & \textbf{98.20\tiny{±0.08}} & \textbf{65.52\tiny{±1.87}} & \textbf{70.79\tiny{±4.17}} & 89.38\tiny{±0.41} & \textbf{77.02\tiny{±7.26}} \\ \hline  
%		\bottomrule\bottomrule
	\end{tabular}
\vspace{-1.0em}
\end{table}

\subsubsection{Scaling Up the Model Capacity}

The last two rows of Table~\ref{table_model_cap} present the results concerning the scaling law of model capacity \cite{scaling}, especially the feature sizes of 3D Swin-B and 3D Swin-Large (L) are 48 and 96, respectively. MG-3D with a larger-scale model (3D Swin-L) can mainly achieve superior performance in vision tasks (the 4th-9th columns) but may experience a decline in cross-modal tasks (the last six columns) . In particular, the obvious performance improvements in Luna16 and MSD Task06 show the potential of larger models to push the achievable performance boundaries.

\vspace{-1.0em}
\subsection{Comparisons with 3D Medical Foundation Models}

We compare the best performance of our VLMs with that of existing SOTA 3D medical FMs. We present the performance of their official models for a comprehensive evaluation. Specifically, we include Vision FMs, such as PCRLv2 \cite{pcrlv2}, Vox2Vec \cite{vox2vec}, TransVW \cite{transvw}, and SwinUNETR \cite{swinunetr}, and VLMs including CT-CLIP \cite{ct_clip}, M3D \cite{m3d}, and RadFM \cite{radfm}.

% PCRLv2 ({\em IEEE T-PAMI} 2023) pre-trained on Luna16 (0.9K) \cite{pcrlv2}, Vox2Vec pre-trained on 6 Mixed Datasets (6.6K) ({\em MICCAI} 2023) \cite{vox2vec}, TransVW pre-trained on Luna16 (0.6K) ({\em IEEE TMI} 2021) \cite{transvw}, SwinUNETR pre-trained on 5 Mixed Datasets (5.1K) ({\em CVPR} 2022) \cite{swinunetr}, and SwinMM pre-trained on 8 Mixed Datasets (5.8K) ({\em MICCAI} 2023) \cite{swinmm}, and VLMs including CT-CLIP pre-trained on CT-RATE (47.1K) ({\em arXiv} 2024) \cite{ct_clip}, M3D pre-trained on M3D-Cap (120K) ({\em arXiv} 2024) \cite{m3d}, and RadFM pre-trained on MedMD (16M) ({\em arXiv} 2024) \cite{radfm}.

\subsubsection{Disease Classification}

As illustrated in the 2nd and 3rd columns of Table~\ref{table_fm_vision}, our pre-trained 3D Swin-L significantly outperform related FMs  on CC-CCII and Luna16, further validating the effectiveness of MG-3D with intra- and inter-patient multi-grained semantics learning.

\subsubsection{Lesion Segmentation}

The visualization results in MSD Task 06 are shown in Fig.~\ref{Fig_Seg}(a)-(c), in which nearly all competitive models have difficulty in precisely segmenting tumors; in contrast, our 3D Swin-L achieves ideal results. Furthermore, Fig.~\ref{Fig_Seg}(d)-(f) show that our 3D Swin-L significantly outperforms related models in segmenting pneumonia with ambiguous boundaries in Covid-19-20. In the 4th and 5th columns of Table~\ref{table_fm_vision}, our VLM also achieves the best performance across all lesion segmentation tasks.

\subsubsection{Organ Segmentation}
The 6th column of Table~\ref{table_fm_vision} indicates that although RadFM demonstrates superior performance due to its pre-training with both CT and MRI images, our 3D Swin-L still showcases comparable cross-modal transferability in 3D MRI image analysis.

\subsubsection{Prognosis Prediction}

Table~\ref{table_fm_vision}'s last column shows that our 3D Swin-L significantly enhances prognosis prediction, owing to its superiority in representing key lesion features.

% \begin{table}[!t]
% \setlength{\tabcolsep}{4pt}
%  \renewcommand{\arraystretch}{1.3}
% 	\caption{Prognosis Prediction Performance of Different 3D Medical FMs on STOIC-2021}
% 	\label{table_fm_prog}
% 	\vspace{-0.5em}
% 	\centering
% 	\begin{tabular}{|c|c|c|}%{|c|l|}
% %		\toprule\toprule
%             \hline
% 		Method & Vision Encoder & AUC (\%) \\ \hline
%             \multicolumn{3}{|c|}{\textbf{{\em With 3D medical SSL}}} \\ \hline
%   		PCRLv2 & 3D UNet & 73.88 \\ \hline
%         Vox2Vec & 3D 3D FPN & 75.27 \\ \hline
%         TransVW & 3D UNet & 56.23 \\ \hline
%         SwinUNETR & 3D Swin-B & 74.46 \\ \hline
%         SwinMM & 3D Swin-B & 71.97 \\ \hline
        
%         \multicolumn{3}{|c|}{\textbf{{\em With Medical VLP}}} \\ \hline
%         CT-CLIP & 3D ViT-B & 63.12 \\ \hline
%         M3D & 3D ViT-B & 62.71 \\ \hline
%         RadFM & 3D ViT-B & 64.85 \\ \hline       
%         \textbf{Ours} & \textbf{3D Swin-L} & \textbf{77.02}\\ \hline  
% %		\bottomrule\bottomrule
% 	\end{tabular}
% \end{table}

% \begin{figure}[!t]
% \centerline{\includegraphics[width=\columnwidth]{Fig_Report.pdf}}
% \setlength{\abovecaptionskip}{-1.0pt}
% \caption{A case study of the report generated by MG-3D, in which the contents with consistent, missing, and similar meanings compared with the ground truth are highlighted in green, red, and orange, respectively. }
% \label{Fig_Report}
% \vspace{-1.5em}
% \end{figure}

\subsubsection{Report Generation}

% \begin{table}[!t]
% \scriptsize
% \setlength{\tabcolsep}{2.5pt}
%  \renewcommand{\arraystretch}{1.3}
% 	\caption{Performance on Lesion Segmentation and Prognosis Prediction of Different 3D Medical FMs on MSD Task 06, Covid-19-20, and STOIC-2021}
% 	\label{table_fm_seg_prog}
% 	\vspace{-0.5em}
% 	\centering
% 	\begin{tabular}{|c|c|c|c|c|}%{|c|l|}
% %		\toprule\toprule
%             \hline
% 		\multirow{2}{*}{\makecell[c]{Method}} & \multirow{2}{*}{\makecell[c]{Vision Encoder}} & MSD Task 06 & Covid-19-20 & STOIC-2021 \\
% 		\cline{3-5}
% 		& & \multicolumn{2}{c|}{Dice (\%)} & AUC (\%) \\ \hline
  
%             \multicolumn{5}{|c|}{\textbf{{\em With 3D Medical SSL}}} \\ \hline
%   		PCRLv2 & 3D UNet & 58.01 & 69.45 & 73.88 \\ \hline
%         Vox2Vec & 3D FPN & 63.33 & 69.77 & 75.27 \\ \hline
%         TransVW & 3D UNet & 51.01 & 69.73 & 56.23 \\ \hline
%         SwinUNETR & 3D Swin-B & 52.83 & 65.84 & 74.46 \\ \hline
%         % SwinMM & 3D Swin-B & 61.23 & 67.20 & 71.97 \\ \hline
        
%         \multicolumn{5}{|c|}{\textbf{{\em With Medical VLP}}} \\ \hline
%         CT-CLIP & 3D ViT-B & 24.93 & 57.16 & 63.12 \\ \hline
%         M3D & 3D ViT-B & 45.07 & 61.39 & 62.71 \\ \hline
%         RadFM & 3D ViT-B & 25.70 & 57.26 & 64.85 \\ \hline
%         \textbf{Ours} & \textbf{3D Swin-L} & \textbf{65.07} & \textbf{70.79} & \textbf{77.02} \\ \hline  
% %		\bottomrule\bottomrule
% 	\end{tabular}
% \vspace{-2.0em}
% \end{table}

The 2nd-5th columns of Table~\ref{table_fm_cross} indicate that our 3D Swin-B owns significant strengthen compared to SOTA FMs, which shows its prospects in advancing 3D radiology semantics understanding. We observe from our generated reports that not only can MG-3D generate reports close to ground truths, but it also effectively aligns different expressions to convey the same underlying representation.

% An example of the generated report is shown in Fig.~\ref{Fig_Report}, in which the main content closely aligns with the ground truth. The descriptions of the pathological conditions and lesion localizations are accurate. Notably, the term "solid nodules" in the generated report corresponds to "spotted high-density shadows" in the ground truth, illustrating that MG-3D effectively aligns different expressions to convey the same underlying representation.

\subsubsection{Vocabulary Prompt-Driven Anomaly Classification \& Report-to-Volume Retrieval}

In the last two columns of Table~\ref{table_fm_cross}, our 3D Swin-B achieves superior performance in anomaly classification and cross-modal retrieval. The large-scale pre-training effectiveness of MG-3D is further verified by the comparisons with CT-CLIP pre-trained based on the same dataset but utilizing a much larger input volume resolution.

\begin{table}[!t]
\scriptsize
\setlength{\tabcolsep}{3.8pt}
 \renewcommand{\arraystretch}{1.3}
	\caption{Performance of Different 3D FMs in Cross-Modal Tasks}
	\label{table_fm_cross}
	\vspace{-0.5em}
	\centering
	\begin{tabular}{|c|c|c|c|c|c|c|}%{|c|l|}
%		\toprule\toprule
            \hline
            
	\multirow{2}{*}{\makecell[c]{Method}} & \multicolumn{4}{c|}{CTRG-Chest} & \multicolumn{2}{c|}{CT-RATE} \\ \cline{2-7}
        & BLEU-1 & BLEU-2 & BLEU-3 & BLEU-4 & AUC (\%)  & Recall@50 \\ \hline
        \multicolumn{7}{|c|}{\textbf{{\em With 3D Medical SSL}}} \\ \hline
  	PCRLv2 & 56.09 & 45.39 & 38.95 & 34.48 & 65.16 & 3.16 \\ \hline
        Vox2Vec & 54.98 & 44.62 & 38.34 & 33.97 & 66.54 & 3.22 \\ \hline
        TransVW & 59.77 & 49.08 & 41.98 & 37.02 & 65.16 & 3.16 \\ \hline
        SwinUNETR & 59.64 & 48.67 & 41.64 & 36.66 & 63.83 & 2.86  \\ \hline
        
        \multicolumn{7}{|c|}{\textbf{{\em With 3D Medical VLP}}} \\ \hline
        CT-CLIP & 59.28 & 48.00 & 40.80 & 35.77 & 66.50 & 3.23 \\ \hline
        M3D & 55.07 & 45.53 & 39.21 & 34.70 & 65.70 & 3.52 \\ \hline
        RadFM & 47.29 & 39.13 & 34.05 & 30.38 & 64.43 & 3.19 \\ \hline     
        \textbf{Ours} & \textbf{63.54} & \textbf{51.61} & \textbf{43.79} & \textbf{38.19} & \textbf{68.62} & \textbf{4.05} \\ \hline  
%		\bottomrule\bottomrule
	\end{tabular}
\vspace{-1.0em}
\end{table}

The comprehensive performance comparisons of different FMs shown in Fig.~\ref{Fig_Radar} reveal the superiority of our VLMs in achieving leading performance across all tasks.

\begin{figure}[!t]
\centerline{\includegraphics[width=\columnwidth]{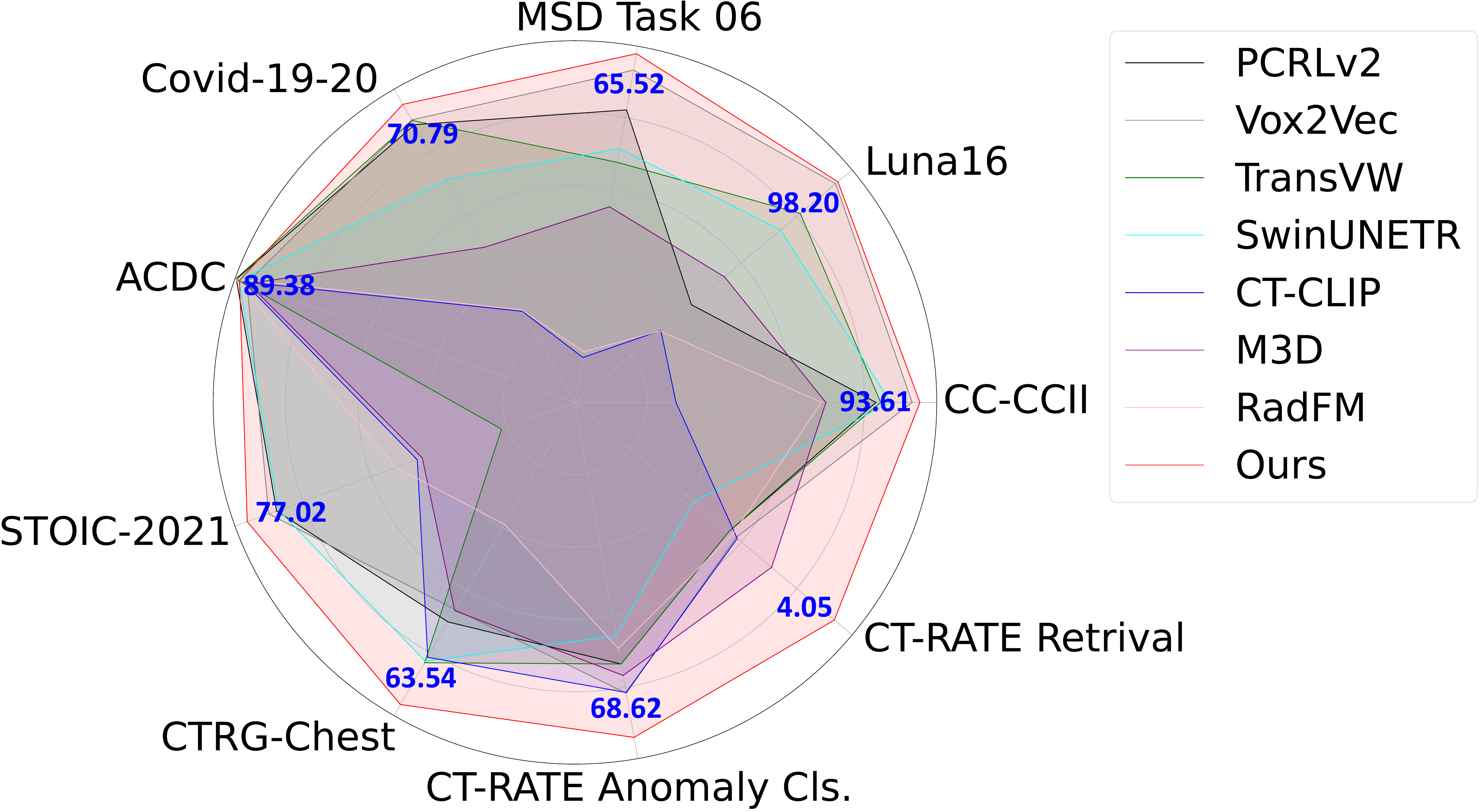}}
\setlength{\abovecaptionskip}{-1.0pt}
\caption{Overall Performance Comparisons of Different Medical FMs. The radar chart clearly shows the superior generalization capabilities of our proposed VLMs.}
\label{Fig_Radar}
\vspace{-1.5em}
\end{figure}

{\vspace{-1.0em}
\section{Conclusion and Future Directions}
\label{sec:conclusion}}

Radiology reports contain detailed descriptions of patients’ anatomical structures and diseases, offering valuable insights for label-efficient representation learning of 3D medical images. To leverage large-scale volume-report data to advance various clinical tasks, we propose a 3D VLP strategy consisting of two main aspects. On one hand, intra-patient multi-grained semantics are effectively extracted by complementary modality-guided local reconstruction and cross-modal global feature alignment with a new cross-modal interaction mechanism, promoting cross-modal semantic understanding. On the other hand, inter-patient multi-grained semantic correlations are captured by inter-patient fine-grained semantics similarity matching and disentangled fine-grained semantics aggregation, strengthening the feature representativeness and discrimination. We also delve into the scaling law of data size and model capacity to unlock potential performance improvements. The effectiveness, scalability, and generalization ability have been extensively evaluated, especially MG-3D achieves the best performance on nine uni- and cross-modal tasks compared with competitive pre-training methods, showcasing the prospect of the proposed VLMs as a foundation to improve the efficiency and adaptability in extensive clinical tasks.

We plan to further collect large-scale 3D volume-report data covering all 3D imaging modalities and full body scenarios, allowing us to develop a more robust, comprehensive, and scalable VLM for advancing 3D medical image analysis. We also found that data heterogeneity poses challenges in 3D VLP, particularly with mixed multi-source data, so we will further explore novel pre-training and data-mixing strategies \cite{doremi}. 

\vspace{0.5em}
\bibliography{bibfile}

\end{document}